\documentclass{article}

\usepackage{microtype}
\usepackage{graphicx}
\usepackage{subfigure}
\usepackage{booktabs} 

\usepackage{hyperref}

\usepackage{algpseudocode}
\usepackage{multirow} 

\usepackage[accepted]{icml2024} 

\usepackage{amsmath}
\usepackage{amssymb}
\usepackage{mathtools}
\usepackage{amsthm}

\usepackage[capitalize,noabbrev]{cleveref}

\theoremstyle{plain}

\theoremstyle{definition}

\theoremstyle{remark}

\usepackage[textsize=tiny]{todonotes}

\begin{document}

\twocolumn[
\icmltitle{Leveraging Image Augmentation for Object Manipulation:\\
Towards Interpretable Controllability in Object-Centric Learning}
\icmlsetsymbol{equal}{*}
\begin{icmlauthorlist}
\icmlauthor{Jinwoo Kim}{equal,yonsei}
\icmlauthor{Janghyuk Choi}{equal,kaist}
\icmlauthor{Jaehyun Kang}{yonsei}
\icmlauthor{Changyeon Lee}{yonsei}
\icmlauthor{Ho-Jin Choi}{kaist}
\icmlauthor{Seon Joo Kim}{yonsei}
\end{icmlauthorlist}
\icmlaffiliation{yonsei}{Yonsei University, South Korea}
\icmlaffiliation{kaist}{Korea Advanced Institute of Science and Technology, South Korea}
\icmlcorrespondingauthor{Seon Joo Kim}{seonjookim@yonsei.ac.kr}
\icmlkeywords{Machine Learning, ICML}
\vskip 0.3in]
\printAffiliationsAndNotice{\icmlEqualContribution} 

\begin{abstract}

The binding problem in artificial neural networks is actively explored to achieve human-level recognition skills through the comprehension of the world in terms of symbol-like entities.
Especially in the field of computer vision, object-centric learning (OCL) is extensively researched to better understand complex scenes by acquiring object representations or \textit{slots}.
While recent studies in OCL have made strides with complex images or videos, the interpretability and interactivity over object representation remain largely uncharted. 
In this paper, we introduce a novel method, Slot Attention with Image Augmentation (SlotAug), to explore the possibility of learning interpretable controllability over slots in a self-supervised manner by utilizing an image augmentation strategy.
We also devise the concept of sustainability in controllable slots by introducing iterative and reversible controls over slots with two proposed submethods: Auxiliary Identity Manipulation and Slot Consistency Loss. 
Extensive empirical studies and theoretical validation confirm the effectiveness of our approach, offering a novel capability for interpretable control of object representations.

\end{abstract}

\section{Introduction}

Compositional comprehension of visual scenes \citep{marr2010vision,johnson2017clevr,fischler1973representation}, essential for various computer vision tasks such as localization \citep{cho2015unsupervised} and reasoning \citep{mao2019neuro}, requires human-like understanding of complex world \citep{treisman1996binding,spelke2007core,lake2017building}.
In response to this, \textit{object-centric learning} (OCL) has emerged as an active research area \citep{locatello2020object,kipf2021conditional,greff2016tagger}.
OCL aims to enable a model to decompose an image into objects, and to acquire their representations, \textit{slots}, without human-annotated labels.

\begin{figure*}[t]
    \centering
    \includegraphics[width=0.8\linewidth]{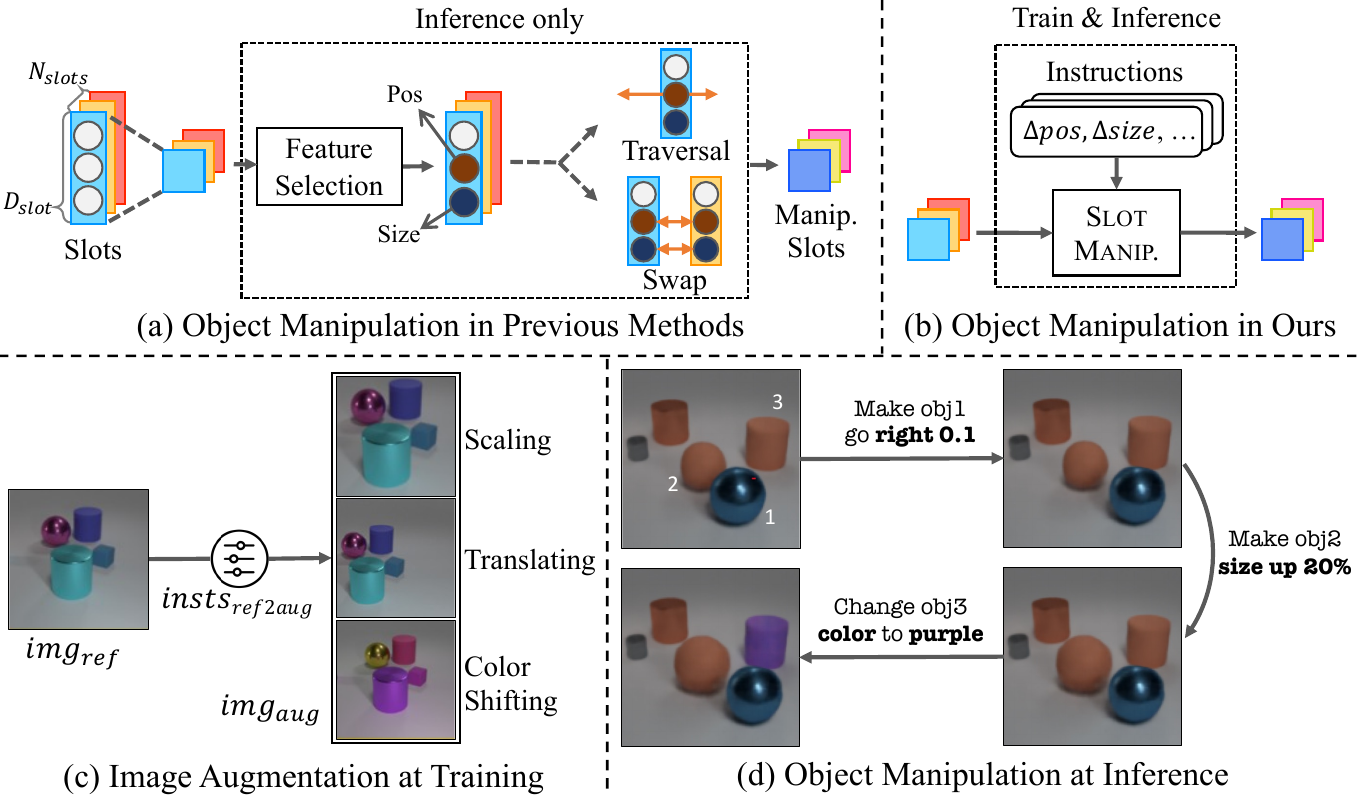}
    \vspace{-1em}
    \caption{
    \textbf{Overview of our method compared to the previous methods.}
    \textbf{(a)} Previous methods require an additional process to manipulate slots such as feature selection during inference. 
    \textbf{(b)} Our model, however, has the shared process of manipulating slots between the training and inference stages.
    \textbf{(c)} To ensure homogeneity between the training and inference stages, we incorporate scenarios involving image manipulation into the training phase. This includes the application of simple image augmentation techniques such as scaling, translating, and color shifting.
    \textbf{(d)} Upon completion of the training, our model achieves interpretable controllability, enabling users to manipulate individual objects according to their intentions.
    }
    \vspace{-1em}
    \label{fig:teaser}
\end{figure*}

In pursuit of a deeper understanding of images, interpretable and controllable object representation has been studied \citep{greff2019multi,burgess2019monet,singh2023neural}.
Nevertheless, the previous approaches face limitations in achieving \textit{interpretable controllability} as they require additional processes to figure out how to interact with slots, such as exploring a connection between specific values in slots and object properties by a manual exhaustive search, and training a feature selector with ground-truth object properties (Fig. \ref{fig:teaser}(a)).
This issue arises due to a training-inference discrepancy, wherein interaction with slots is only considered during the inference stage.
This discrepancy brings ambiguity in how to interact with object representations, hindering interpretable controllability.
Furthermore, learning interpretable controllability is followed by another challenge: object representation should be intact even after multiple manipulations by humans.
In this context, we devise the concept of \textit{sustainability} pertaining to the ability to preserve the nature of slots, allowing for iterative manipulations; we refer to Fig. \ref{fig:exp_accum} and \ref{fig:exp_durability} for establishing the earlier motivation.

In this work, we advance the field of OCL with respect to the interpretability of object representation.
To achieve interpretable controllability, we propose a method that enables the manipulation of object representation through human interpretable instructions in a self-supervised manner.
We address the training-inference discrepancy by incorporating image augmentation into our training pipeline (Fig. \ref{fig:teaser}(c)).
By involving the slot manipulation in the training, we can resolve the discrepancy and streamline the way to interact with slots in the inference stage (Fig. \ref{fig:teaser}(b) and (d)). 

Second, to attain sustainability in object representation, we introduce \textit{Auxiliary Identity Manipulation} (AIM) and \textit{Slot Consistency Loss} (SCLoss).
AIM is a methodology designed to facilitate the learning of the concept of multi-round manipulation. 
AIM is implemented by incorporating an auxiliary manipulation process into the intermediate stage of slot manipulation, where the auxiliary manipulation introduces no semantic changes to object properties such as zero-pixel translations.
This simple auxiliary process can expose our model to multi-round manipulation: we can make two-round manipulations with one instruction from the augmentation and the other from the auxiliary manipulation.
Additionally, SCLoss allows our model to learn the concept of reversible manipulation, such as the relationship between moving an object 1 pixel to the right and moving it 1 pixel to the left.
After being trained with SCLoss, our model produces consistent and reusable representations that can undergo multiple modifications and enhance their usability.
With AIM and SCLoss, our model achieves sustainability in object representation.

Extensive experiments are shown to demonstrate the interpretable and sustainable controllability of our model.
To assess interpretability, we conduct object manipulation experiments where slots are guided by semantically interpretable instructions. 
In evaluating sustainability, we introduce novel experiments, including the durability test. 
Our evaluation encompasses not only pixel space assessments such as image editing via object manipulation, but also slot space analyses such as property prediction, to provide a comprehensive examination of our approach.


\section{Methods}

\begin{figure*}
    \centering
    \includegraphics[width=0.9\linewidth]{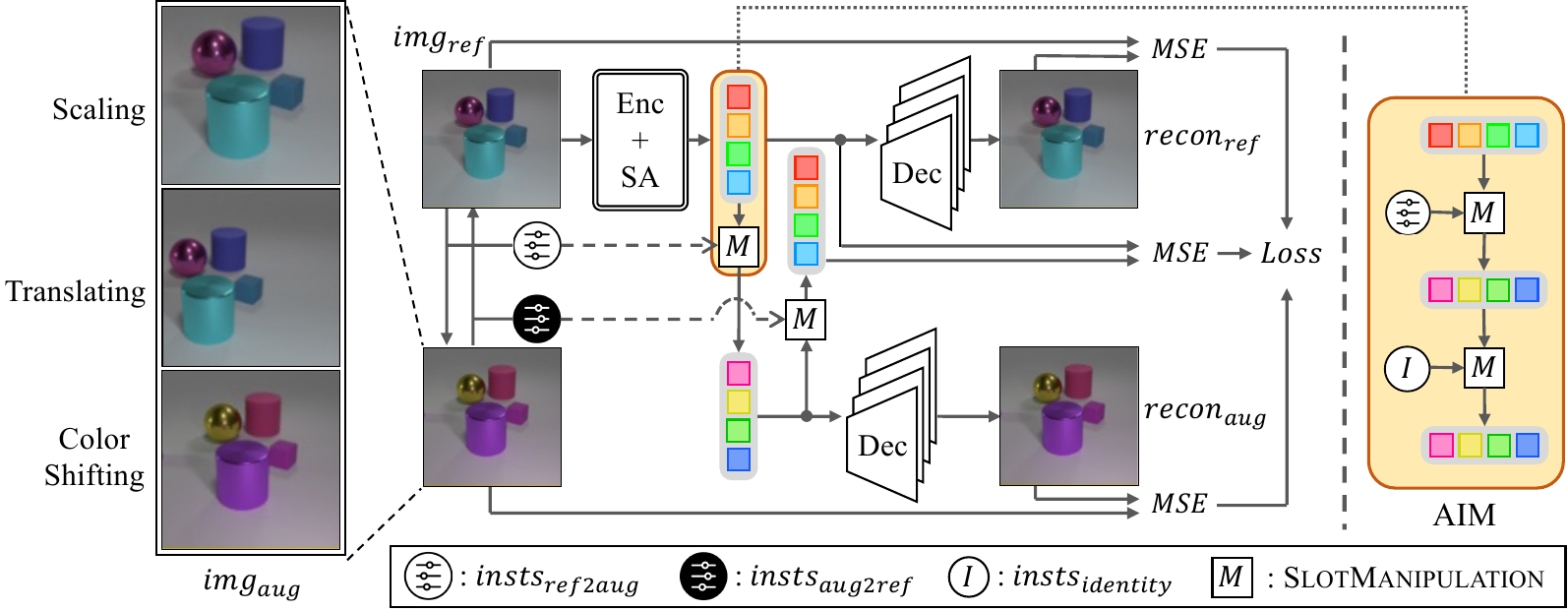}
    \vspace{-1em}
    \caption{
    \textbf{Architecture of our model.}
    From a given image $img_{ref}$, we first generate an augmented image $img_{aug}$ (leftmost part of the figure), and the corresponding instruction $insts_{ref2aug}$ and its inverse $insts_{aug2ref}$. 
    Our model produces slots from $img_{ref}$ and decodes the slots to reconstruct the original image ($recon_{ref}$).
    The slots are also manipulated with \textsc{SlotManipulation} module which takes $insts_{ref2aug}$ as the other input.
    We incorporate Auxiliary Identity Manipulation (AIM) into this manipulation process. The details are provided in the right part of the figure.
    The manipulated slots are then simultaneously 1) decoded into a reconstruction of the augmented image $recon_{aug}$, and 2) re-manipulated by \textsc{SlotManipulation} with $insts_{aug2ref}$.
    Our total loss consists of the reconstruction losses of reference and augmented images, and the slot-level cycle consistency.
    }
    \label{fig:model}
\end{figure*}

\subsection{Preliminary: Slot Attention}
Slot Attention (SA) \citep{locatello2020object} introduces the concept of \textit{slots}, a set of $K$ vectors of dimension $D_{slot}$, that serves as the object representation.
The slots are initialized by a Gaussian distribution with learnable mean $\mu$ and sigma $\sigma$, and are updated over $T$ iterations by the slot attention module.
The final slots are then decoded to reconstruct the target image.
To provide a comprehensive understanding of our method, we describe the mechanism of spatial binding \citep{greff2020binding,treisman1996binding,buehner2010causal} by Slot Attention in Alg. 2, referred to as \textsc{SpatialBinding}, in the Appendix due to the space limitation.
Each updated slot is then decoded individually into an RGBA image using a spatial broadcast decoder \citep{watters2019spatial} which is shared across slots. 
The decoded images are blended into a single image using alpha masks to reconstruct the input image.
The training objective is the mean squared error (MSE) between the original input image and the reconstructed image, following a self-supervised learning approach.

\subsection{SlotAug: Slot Attention with Image Augmentation}
\label{subsec:interpretable_controllability}

\noindent \textbf{Data augmentation.}
We introduce a simple data augmentation scheme that, for a given input reference image $img_{ref} \in \mathbb{R}^{H \times W \times 3}$, generates an augmented image $img_{aug} \in \mathbb{R}^{H \times W \times 3}$ and the transformation instructions between them, $insts_{ref2aug} \in \mathbb{R}^{K \times L}$, where $L$ indicates the number of values to represent the object properties.
$img_{aug}$ is produced by a random translation, scaling, or color shifting on $img_{ref}$.
To transform $img_{ref}$ into $img_{aug}$, we employ a set of instructions known as $insts_{ref2aug}$. 
These instructions comprise a list of values that dictate the augmentation process, including translation values, a scaling factor, and color shift values in the HSL color space. 
We also have the inverse instructions, $insts_{aug2ref} \in \mathbb{R}^{K \times L}$, which allow us to revert $img_{aug}$ back to $img_{ref}$.

Henceforth, for the sake of simplicity in notation, we will employ $r$ and $a$ as shorthand for $ref$ and $aug$. 
For instance, the expression $img_{r}$ and $img_{r2a}$ will be used to denote $img_{ref}$ and $img_{ref2aug}$, respectively.
More details are described in the Appendix.

\noindent \textbf{Training.}
We propose a novel training process that leverages image augmentation (Fig. \ref{fig:model}).
Our training scheme enables learning interpretable controllability which allows us to interact with the model via semantically interpretable instructions.
Our training process involves data augmentation, spatial binding, slot manipulation, and image reconstruction via slot decoding.
For a given input image, we initially perform data augmentation to yield $img_{r}$, $img_{a}$, $insts_{r2a}$, and $insts_{a2r}$. 
Then, the model performs \textsc{SpatialBinding} on $img_{r}$ to produce $slots_{ref}$, or $slots_{r}$.

Thereafter, the model conducts \textsc{SlotManip} (Alg. \ref{alg:slotaug}) to modify $slots_{r}$ based on $insts_{r2a}$.
In the \textsc{SlotManip}, we utilize a newly introduced component called \textit{PropEnc}, which is 3-layer multi-layer perceptrons (MLPs). 
This PropEnc generates vector representations, $\texttt{inst}\_\texttt{vec}$, which capture the essence of transformation instructions.
Each $\text{PropEnc}_{j}$ generates an $\texttt{inst}\_\texttt{vec}_\texttt{j}$ that encodes the values of $insts_{r2a}$ for the $j$-th property.
These vectors are then added to $slots_{r}$ to reflect the effect of $insts_{r2a}$.
This addition is followed by a residual connection, along with layer normalization and another MLP to generate $slots_{r2a}$.

Lastly, $slots_{r2a}$ is decoded by the decoder to create the $recon_{a}$, the reconstruction for the augmented image $img_{a}$.
The MSE between $img_{a}$ and $recon_{a}$ serves as a training loss, $\mathcal{L}_{\text{aug}}$.
To ensure stable training, we also adopt an additional loss, $\mathcal{L}_{\text{ref}}$, the MSE between the $img_{r}$ and $recon_{r}$, the reconstructed reference image decoded from $slots_{r}$.
Accordingly, our training loss for image reconstruction is defined as $ \mathcal{L}_{\text{recon}} = \mathcal{L}_{\text{ref}} + \mathcal{L}_{\text{aug}}$.

\begin{algorithm}[t]
	\caption{
 Our slot manipulation algorithm in pseudo code.
 \textit{J} represents the number of object properties, while $\texttt{P}_{j,f}$ and $\texttt{P}_{j,l}$ indicate the first and last indices of the j-th object property values. 
 The algorithm uses Layer Normalization \texttt{LN} to normalize vectors.
 } 
	\begin{algorithmic}[1]
        \Function{\text{SlotManip}}{$\texttt{slots}, \texttt{insts}$}
            \For{$j = 0 \ldots J$} 
                \State $\texttt{inst}_{j} = \texttt{insts}[:, \texttt{P}_{j,f}: \texttt{P}_{j,l}]$
                \State $\texttt{inst\_vec}_{j} = \texttt{PropEnc}_{j}(\texttt{LN}(\texttt{inst}_{j}))$
                \State $\texttt{slots} = \texttt{slots} + \texttt{inst\_vec}_{j}$
            \EndFor
        \State $\texttt{slots} = \texttt{slots} + \texttt{MLP(LN(slots))}$
        \State \textbf{return} \texttt{slots}
        \EndFunction
	\end{algorithmic} 
\label{alg:slotaug}
\end{algorithm}


\noindent \textbf{Inference.}
To perform object manipulation, we provide the model with the position of the target object, along with the instruction to be carried out.
When given the position of the target object, we use the Hungarian algorithm \citep{kuhn1955hungarian} to find the slot for the object closest to the given position.
To predict the position of an object encoded in a slot, we compute the center of mass acquired from the alpha mask by the decoder or from the attention map between the visual encodings and the slot.
After figuring out the desired slot, we perform slot manipulation with the given instructions.

\subsection{Sustainability in Object Representation}
In this work, we introduce \textit{sustainability} which stands for the concept that object representations should sustain their integrity even after undergoing iterative manipulations.
Therefore, sustainability is a key feature that contributes to the reliable and reusable object representation.

\noindent \textbf{Auxiliary Identity Manipulation (AIM)} serves as the identity operation for slot manipulation, indicating no changes in object properties. 
By manipulating slots with instructions that include zero values for translation, one for scaling, and so on, AIM is supposed to make each slot preserve the original identity of the object.
We incorporate AIM into the training process to make the model recognize and maintain the intrinsic characteristics of objects during iterative manipulations.
AIM is applied to the slot manipulation process as follows:
\begin{equation}
    \begin{split}
    slots'_{r2a} = f(f(slots_{r}, insts_{r2a}), insts_{id}) \\
    = f(slots_{r2a}, insts_{id}), \phantom{00000000}
    \end{split}
    \label{eq:aux_identity_manip}
\end{equation}
where $f$ represents the \textsc{SlotManip} function, and $insts_{id}$, or $insts_{identity}$, is the instruction that contains the identity elements for manipulating properties.
In the followings, $slots'_{r2a}$ is notated as $slots_{r2a}$ for simplicity. 

\noindent \textbf{Slot Consistency Loss (SCLoss)} addresses the issue of a slot diverging significantly from its original state after iterative manipulations, even when a user intends to restore the corresponding object to its original state.
To implement SCLoss, we introduce $slots_{restored}$, which is derived by executing a series of \textsc{SlotManip} operations on $slots_{r}$ using $insts_{r2a}$ and $insts_{a2r}$.
Supposed that our goal is to ensure $slots_{r}$ and $slots_{restored}$ have the same representation, we set the MSE between them as SCLoss.
As a result, the model learns to keep the two distinct slots representing the same object as close as possible and to be robust against multiple rounds of manipulation.
The equation of SCLoss, $\mathcal{L}_{\text{cycle}}$, and the total training loss, $\mathcal{L}_{\text{total}}$, are as follows:


\begin{equation}
    \mathcal{L}_{\text{cycle}} = \frac{1}{K} \| f(slots_{r2a}, insts_{a2r}) \\
    - slots_{r} \|_2^2,
    \label{eq:cycle_loss}
\end{equation}
\begin{equation}
    \mathcal{L}_{\text{total}} = w_{recon} \mathcal{L}_{\text{recon}} + w_{cycle} \mathcal{L}_{\text{cycle}},
\end{equation}
where $K$ is the number of slots, $f$ is the \textsc{SlotManip} function, and
$w_{recon}$ and $w_{cycle}$ are the weights for the corresponding loss.

\section{Related Works}

The binding problem in artificial neural networks \citep{greff2020binding}, inspired by cognitive science \citep{treisman1996binding,feldman2013neural}, is a subject of active exploration, aiming to attain human-like recognition abilities by understanding the world in terms of symbol-like entities such as objects.
In computer vision, object-centric learning (OCL) focuses on comprehending visual scenes by considering objects and their relationships without labeled data \citep{xie2022coat, engelcke2021genesis,wu2022slotformer}. 
MONet \citep{burgess2019monet}, IODINE \citep{greff2019multi}, and GENESIS \citep{engelcke2019genesis} have adopted autoencoding architectures \citep{baldi2012autoencoders,kingma2013auto,makhzani2015adversarial} to accomplish self-supervised OCL, and Slot Attention \citep{locatello2020object} introduced the concept of slot competition, which enables parallel updates of slots with a single visual encoding and decoding stage. 
Recent studies have leveraged large-scale models to learn object representations in complex images \citep{singh2021illiterate, seitzer2022bridging}, multi-view images \citep{sajjadi2022object}, and videos \citep{kipf2021conditional, singh2022simple}.
Other recent works have utilized object-related inductive biases to improve the OCL models. 
SLASH \citep{kim2023shepherding} addressed the instability in background separation using a learnable low-pass filter to solidify the object patterns in the attention maps. 
SysBinder \citep{singh2023neural} introduced a factor-level slot, called block, to disentangle object properties and enhance the interpretability in OCL.

Several studies have shown the possibility of interacting with object representation to manipulate the objects.
VAE-based models such as IODINE \citep{greff2019multi} and Slot-VAE \citep{wang2023slot} showed that adjusting the values of slots can change object properties. 
SysBinder \citep{singh2023neural} demonstrated that replacing factor-level slot, called block, between slots exchanges the corresponding properties.
However, these works have difficulties in determining ways to interact with slots as they require manual efforts to identify the features associated with specific properties. 
ISA \citep{biza2023invariant} incorporates spatial symmetries of objects using slot-centric reference frames into the spatial binding process, enhancing interactivity of object representation for spatial properties such as position and scale.
Meanwhile, our method itself has no constraint on the types of the target property, showing its expandability toward extrinsic properties such as the shape and material of objects if there exist proper image augmentation skills or labeled data. 
In another direction, MulMon \citep{li2020learning} and COLF \citep{smith2022unsupervised} showed the manipulation of extrinsic object properties, such as position and z-axis rotation, by utilizing a novel view synthesis with a multi-view dataset. 
In contrast, our work accomplishes direct and interpretable controllability over object representation in single-view images without requiring multi-view datasets.

\begin{figure*}
    \centering
    \includegraphics[width=\linewidth]{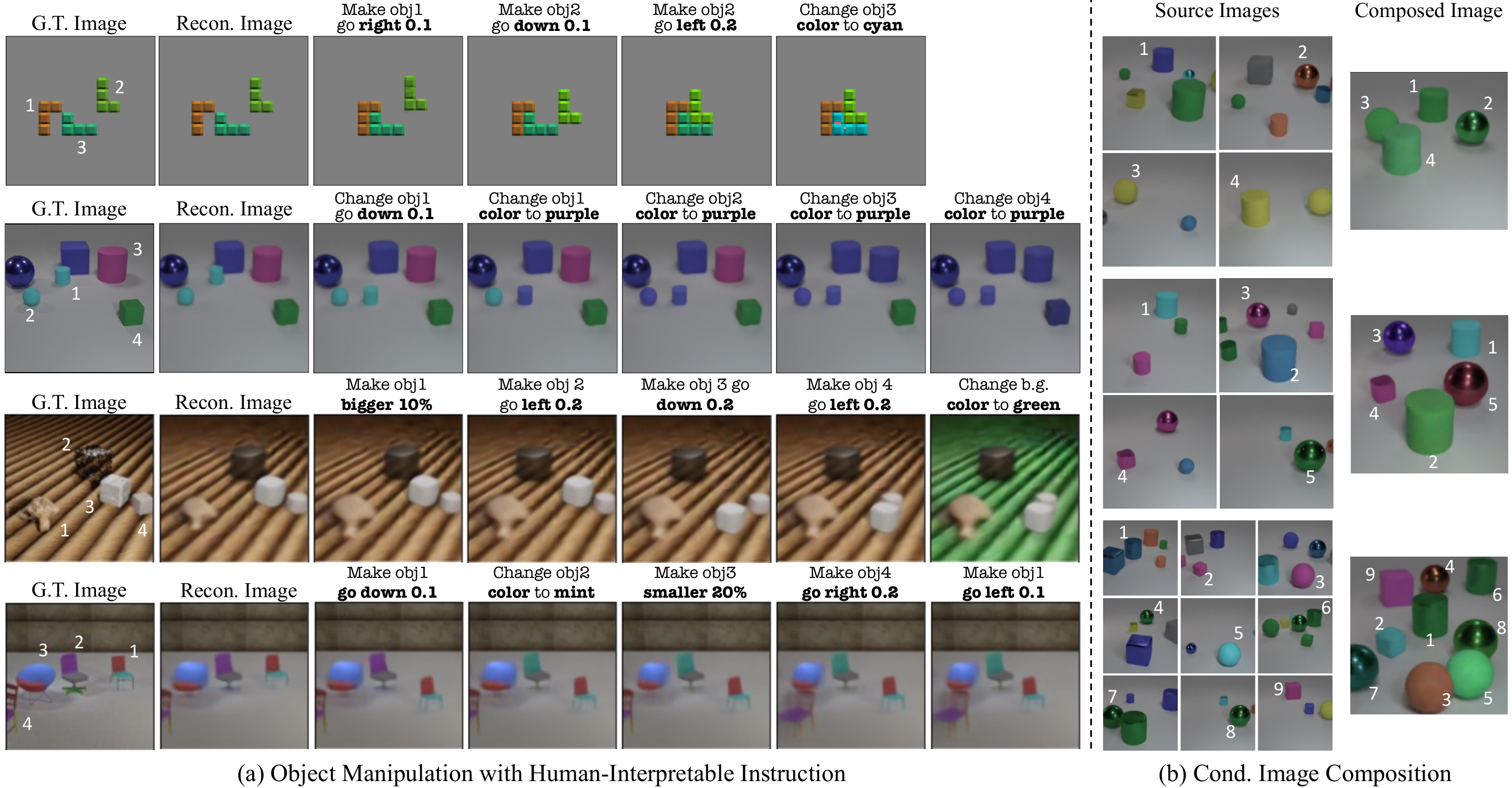}
    \vspace{-1em}
    \caption{
        \textbf{(a) Object manipulation with human-interpretable instruction.}
        The first and second columns are the ground-truth and reconstruction images, respectively.
        The following columns are the results of the controls along the instructions.
        Here, instructions are described with the text for easy understanding.
        The actual instantiation of the instructions can be found in the Appendix.
        From the first row onwards, the results are for Tetrominoes, CLEVR, CLEVR, and PTR, respectively.
        \textbf{(b) Conditional image composition.}
        From given source images, we can collect specific objects, which are indicated by white numbers, and manipulate them to generate a novel image.
    }
    \label{fig:exp_manip_new}
\end{figure*}

\section{Experiments}
\label{sec:experiments}

\noindent \textbf{Datasets.}
We evaluate models on four multi-object datasets: Tetrominoes \citep{multi-object-datasets}, CLEVR6 \citep{johnson2017clevr}, CLEVRTEX6 \citep{karazija2021clevrtex} and PTR \citep{hong2021ptr}.
For Tetrominoes, we use 60K and 15K samples as a train and a test set, respectively.
CLEVR6 is a subset of CLEVR dataset, where $6$ stands for the maximum number of objects in a scene.
We use 35K samples for training and 7.5K samples for testing.
CLEVRTEX6 constitutes a subset of CLEVRTEX, which, in turn, represents a more complex variation of CLEVR. 
CLEVRTEX introduces difficulty through intricate shapes, textures, materials, and backgrounds.
CLEVRTEX6 contains 20K and 5K samples for training and testing.
PTR is a dataset consisting of 52K training and 9K test samples, containing sharp and structured objects with part-whole hierarchies, posing challenges for comprehension, particularly in a self-supervised setting.


\noindent \textbf{Training.}
Unless stated otherwise, the training setup follows the methodology in Slot Attention \citep{locatello2020object}.
The number of epochs is 1000 with 20 warm-ups and 200 decaying epochs.
We adopt AdamW \citep{loshchilov2019decoupled} as the optimizer. 
The number of slots ($K$) is set to 7 and the input image size ($H \times W)$ is set to 128 $\times$ 128, except for Tetrominoes where $K = 7$ and $H=W=64$.
The weights for the training loss are set as $w_{recon} = 1.0$ and $w_{cycle} = 0.1$.
The details of the training process including the data augmentation setting are stated in the Appendix.

\noindent \textbf{Models.}
Basically, we employ the model architecture of Slot Attention.
For CLEVRTEX6 and PTR, we replace the encoder with ViT \citep{dosovitskiy2020image} pretrained by MAE \citep{he2022masked} and the decoder with that of SRT \citep{sajjadi2022scene} while using an increased size of the slot attention module.
The additional details for adopting large models are described in the Appendix.
To clarify the methods used in ablative studies, we categorize our model into three versions: v1, which is exclusively trained with image augmentation; v2, which improves upon v1 with AIM; v3, which extends v1 with both AIM and SCLoss. 
For qualitative studies, we use the v3 model.

\subsection{Interpretable Control over Object Representation}
\subsubsection{Image Editing by Object Manipulation} 
As shown in Fig. \ref{fig:exp_manip_new}(a), our model can manipulate individual objects.
We can apply control over multiple objects and various properties of a single object, and even manipulate a single object multiple times, all in accordance with the specific intentions conveyed through user instructions.
Based on this observation, we can ascertain that our object representations retain the intrinsic properties of objects seamlessly even after manipulation.
This interpretable controllability is accomplished with a neglectable compromise of the performance on the object discovery task, as demonstrated in Tab. \ref{tab:od_and_recon}.
While not minor, it is worth noting that we can also manipulate the background alongside objects as shown in the third row of Figure \ref{fig:exp_manip_new}(a) since our model incorporates the SLASH architecture \cite{kim2023shepherding} which treats the background as a single entity.

One may wonder how our model can excel in controlling individual slots while being trained solely on image-level manipulation without any explicit object-level supervision.
We attribute this successful transition from image augmentation to object manipulation to the slots' ability to focus their attention effectively on each distinct object.
This capability is achieved by the following key factors. 
Firstly, using slot-wise decoder \citep{locatello2020object} enables independent decoding for each slot, eliminating dependencies on other slots. 
Secondly, using Attention Refining Kernel (ARK) \citep{kim2023shepherding} allows our model to efficiently discover individual objects without concerns of attention leakage.
These factors collectively ensure that the slots remain directed toward their corresponding objects, facilitating precise object manipulations.
We claim that these factors enable our model to seamlessly extend the knowledge learned from \textit{image} augmentation to \textit{object} manipulation.
Deeper discussions with theoretical proof and empirical results are provided in the Appendix to substantiate our claim.

\subsubsection{Conditional Image Composition}
We introduce \textit{conditional image composition}, an advanced version of a downstream task called compositional generation \citep{singh2021illiterate}.
From compositional generation, we can evaluate the reusability and robustness of slots obtained from different images.
As shown in Fig. \ref{fig:exp_manip_new}(b), our task aims to generate novel images by not only combining but also \textit{manipulating} slots from various images. 

Owing to the direct controllability, our model is capable of rendering objects along with human intention by modifying the objects in accordance with instructions that contain values for the desired change.
As shown in the row of Fig. \ref{fig:exp_manip_new}(b), the number of slots for image composition (9 objects) can be expanded beyond the quantity for which the model was originally trained (up to 6 objects).
We attribute this to the characteristics of slots in Slot Attention.
Moreover, our model also can resolve the conflicts among multiple images regarding the relative position (or depth) of the objects as illustrated in the object 3, 5 and 8 in the third row in Fig. \ref{fig:exp_manip_new}(b).
From these observations, we claim that the proposed method can effectively manipulate and combine slots without sacrificing the original nature and robustness.

\begin{table}
\centering
\caption{\textbf{Results on object discovery.} Two metrics, mean Intersection over Union (mIoU) and Adjusted Rand Index (ARI) are reported in \% (mean $\pm$ std for 3 trials) on CLEVR6. }
\label{tab:od_and_recon}
\resizebox{0.9\linewidth}{!}{%
\begin{tabular}{lcc}
\toprule
 & mIoU ($\uparrow$) & ARI ($\uparrow$) \\ \hline
 \multicolumn{1}{c}{Previous methods} & &  \\ \hline
SA \citep{locatello2020object} & 47.3 $\pm$ 23.2 & 63.1 $\pm$ 54.5 \\
+ARK \citep{kim2023shepherding} & \textbf{68.8 $\pm$ \phantom{0}0.4}  & \textbf{95.4 $\pm$ \phantom{0}0.5} \\ \hline
\multicolumn{1}{c}{SlotAug (Ours)} & & \\ \hline
(v1) Train w/ aug. & \textbf{68.9 $\pm$ \phantom{0}0.1} & \textbf{95.7 $\pm$ \phantom{0}0.2} \\
(v2) + AIM & 68.5 $\pm$ \phantom{0}0.1 & 95.3 $\pm$ \phantom{0}0.1 \\
(v3) + AIM + SCLoss & 68.5 $\pm$ \phantom{0}0.1 & \textbf{95.2 $\pm$ \phantom{0}0.7} \\
\bottomrule
\end{tabular}%
}
\end{table}

\subsection{Sustainability in Object Representation}

\begin{figure*}[t]
    \centering
    \includegraphics[width=0.85\linewidth]{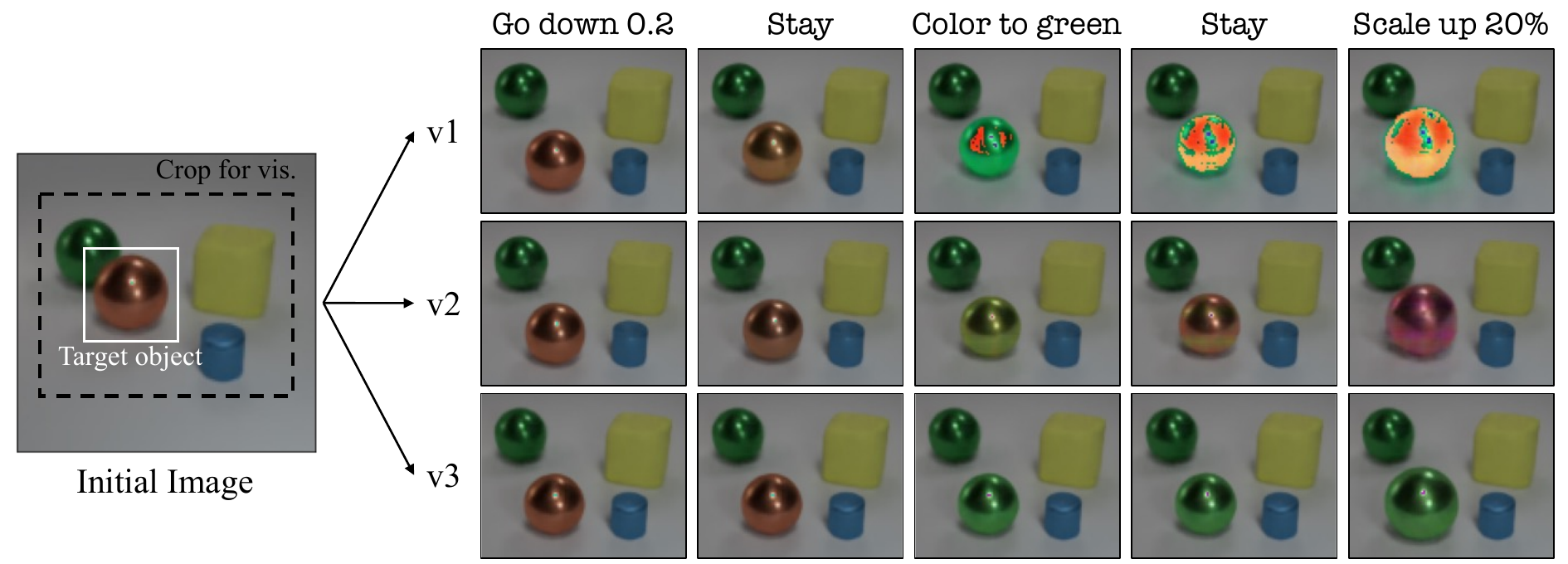}
    \vspace{-0.7em}
    \caption{
    \textbf{Iterative slot manipulation. }
    The leftmost image is the initial image from which the iterative manipulation begins. 
    The text on each column states the instruction used for manipulation.
    Each row shows the results of the manipulation by v1, v2, and v3 models, respectively.
    Center areas are cropped for better visibility.
    }
    \label{fig:exp_accum}
\end{figure*}

\begin{figure*}
    \centering
    \includegraphics[width=0.85\linewidth]{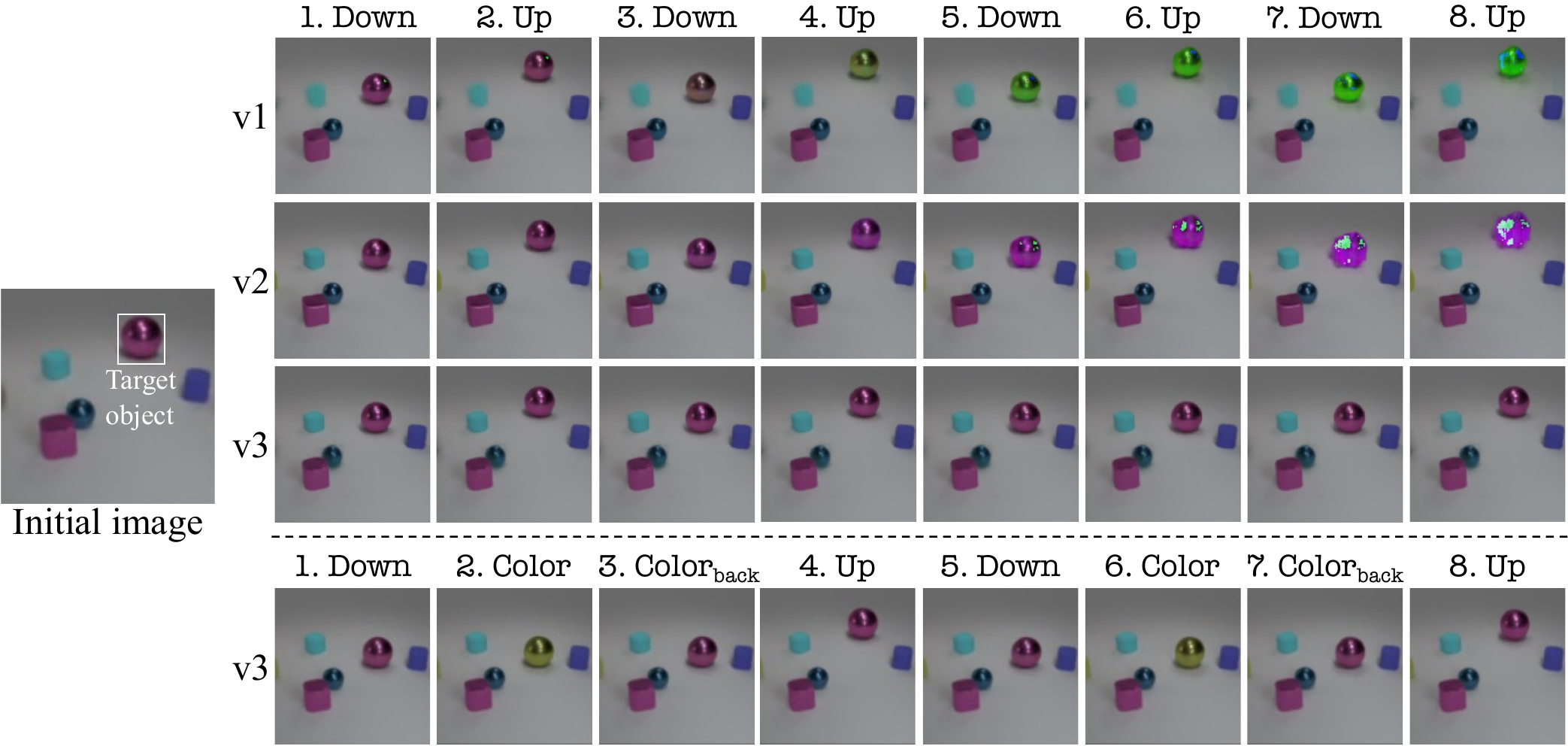}
    \vspace{-0.5em}
    \caption{
    \textbf{Durability test. }
    The leftmost image is the initial image from which the test begins. 
    The top three rows show the results of the single-step tests where each model is instructed to alternately move the target object up and down four times each.
    In the multi-step test, as shown in the last row, the model performs two round-trip manipulations, each involving moving the target object down, changing its color, reverting the color, and returning the object to its original position.
    }
    \label{fig:exp_durability}
\end{figure*}

\subsubsection{Iterative manipulation}
\label{subsubsec:multi_manip}
Fig. \ref{fig:exp_accum} shows the results of iterative manipulation applied to a specific object along with a series of instructions including "Stay" referring to $insts_{id}$.
We can observe that all our models succeed in manipulating the target object, demonstrating that our proposed training scheme works properly.
Nevertheless, it is also clear that models v1 and v2 fail to follow the instructions along the consecutive manipulations. 
In the case of v1, object appearances deteriorate with the emergence of abnormal artifacts from the third round.
Whereas, in v2, although the collapsing issue is mitigated, an out-of-interest property, color, changes despite no instruction for such modification.
These unexpected results are also triggered by the "Stay" instruction, which is intended to maintain the current state of the object.
However, in the case of v3, we finally achieve optimal results that adhere to the instructions, including "Stay".
Based on these observations, we argue that both AIM and SCLoss significantly contribute to sustainable controllability.

\subsubsection{Durabiltiy test}

In the durability test, we evaluate how many manipulations a model can endure while preserving object representation intact. 
Our durability test consists of two types: single- and multi-step tests.
In the single-step test, we repeatedly manipulate slots with two instructions: one to modify a specific object property and another to revert the object to its initial state.
The multi-step test involves a series of instructions to modify an object and another series to restore it to its initial state.

As depicted in Fig. \ref{fig:exp_durability}, our findings align with the previous experiment (Sec. \ref{subsubsec:multi_manip}).
While v1 fails to keep the color after the second round and the color gradually deviates as the rounds progress, v2 relatively preserves the color well for the fourth round.
Nevertheless, from the fifth round, the texture progressively diverges from its original.
Different from the v1 and v2, v3 demonstrates strong durability despite a greater number of manipulations.

We also perform quantitative evaluations on 100 randomly selected samples from CLEVR6 to measure the intrinsic deformity of slots and extrinsic change of object properties, especially position (Tab. \ref{tab:durability}).
We conduct 8 single step and 4 multiple steps round trip manipulations, both resulting in a total of 16 manipulations.
We assess the durability test results by measuring the difference, using L2 distance, between the original state and the manipulated state for two aspects: the slot vector and object position vector.
From both qualitative and quantitative results, we can easily tell that our model can achieve better sustainability as the model evolves from v1 to v2 and v3 by utilizing the proposed AIM and SCLoss.
More extreme durability tests are illustrated in the Appendix.

\begin{table}
\centering
\caption{
\textbf{Results on durability test} with MSE on CLEVR6.
}
\label{tab:durability}
\resizebox{.8\linewidth}{!}{%
\begin{tabular}{lcc}
\toprule
 & Slot ($\downarrow$) & Obj. Pos. ($\downarrow$) \\ \hline
 & \multicolumn{2}{c}{Single step (x8)} \\ \hline
(v1) Train w/ aug. & 50.8 & 0.14 \\
(v2) + AIM & 39.7 & 0.15 \\
(v3) + AIM + SCLoss & \textbf{0.25} & \textbf{0.01} \\ \hline
 & \multicolumn{2}{c}{Multiple steps (x4)} \\ \hline
(v1) Train w/ aug. & 54.0 & 0.16 \\
(v2) + AIM & 41.4 & 0.11 \\
(v3) + AIM + SCLoss & \textbf{0.31} & \textbf{0.02} \\
\bottomrule
\end{tabular}
}

\end{table}

\begin{figure*}
    \centering
    \includegraphics[width=\linewidth]{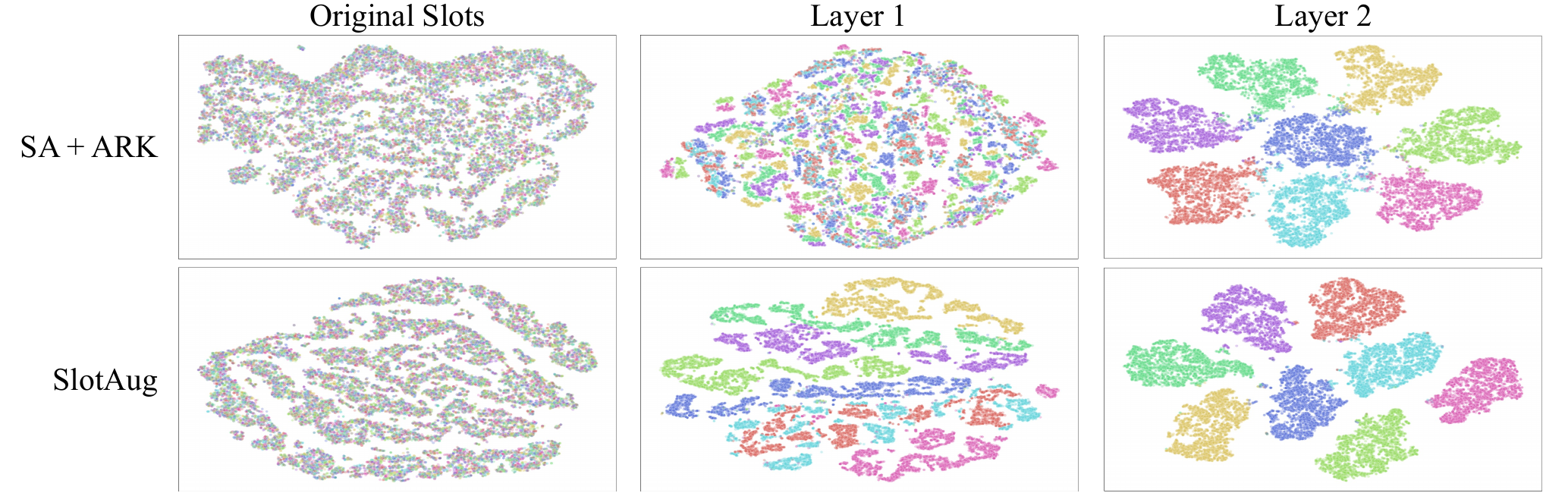}
    \vspace{-0.7em}
    \caption{
    \textbf{t-SNE of slots on property prediction for color.} 
    The upper row is the results for the baseline model, SA + ARK, and the lower row is the results for our method, SlotAug.
    The first column is the result of the original slots obtained from the spatial binding process. 
    The second and third columns are the results of the intermediate outputs from the first and second MLP layers of the property predictor, respectively.
    The results of t-SNE for other properties are shown in the Appendix.
    }
    \label{fig:tsne}
\end{figure*}

\subsection{Slot Space Analysis: Property Prediction}


 In addition to the pixel space analysis represented in the previous sections, for comprehensive assessment of the effectiveness of our method, we further extend our examination to the analysis of the latent slot space. 
To evaluate the quality of slots concerning human-interpretable object properties, such as size, color, material, shape, and position, we conduct a \textit{property prediction} task using the CLEVR6 dataset.
This task enables us to scrutinize how well the slots are distributed within the latent vector space in accordance with the properties of corresponding objects.  
The analysis allows for an understanding of object representations that impart semantic existence significance beyond mere binding or segmenting of objects in an image, offering a deeper insight into the representational aspects of slots. 

A \textit{property predictor}, consisting of 3-layer MLPs, takes slots as input and predicts a property of objects.
Each property predictor is trained by supervised learning using the ground truths matched by the Hungarian algorithm \citep{kuhn1955hungarian}.
To investigate the effectiveness of object representations learned through the proposed method, we freeze the OCL models that produce slots during property prediction.
As shown in Tab. \ref{tab:pp}, our model outperforms the baseline method \citep{kim2023shepherding} across all properties including those, like material and shape, that are not addressed during training.
Moreover, in Fig. \ref{fig:tsne}, qualitative results using t-SNE \citep{van2008visualizing} show that while the original slots themselves do not appear to be well-clustered, slots obtained by SlotAug exhibit better adaptability to the downstream task compared to those from the baseline model.
This observation reinforces our quantitative findings.
Based on these results, we assert that our method enhances interpretability not only in the pixel space but also in the slot space.

\begin{table}[t]
\centering
\caption{\textbf{
    Results of property prediction on CLEVR6.} 
    Each column reports the F1 score (\%) for predicting size, color, material, shape, and position, respectively.
    The numbers inside the parenthesis indicate the number of classes.
    For the position, we set two distance thresholds indicated as `pos@threshold'.
}
\label{tab:pp}
\resizebox{.8\linewidth}{!}{%
\begin{tabular}{lcc}
\toprule
 & SA + ARK & SlotAug (Ours) \\ \hline
size (2) & 69.7 & \textbf{82.2} \\
color (8) & 63.5 & \textbf{78.2} \\
material (2) & 70.4 & \textbf{82.6} \\
shape (3) & 59.1 & \textbf{73.0} \\
pos@0.15 & 71.1 & \textbf{84.2} \\
pos@0.05 & 51.8 & \textbf{77.2} \\
\bottomrule
\end{tabular}%
}
\vspace{-1.0em}
\end{table}


\section{Conclusion}

We presented an OCL framework, SlotAug, for exploring the potential of interpretable controllability in object-centric learning.
To achieve this goal, we tackled the object manipulation task, where we added some conditions regarding interpretability and interactivity, via controlling object representations called slots.
We employed image augmentation for training our model in a self-supervised manner to resolve the lack of labeled data.
Moreover, we introduced a concept of sustainability in slots, achieved by the proposed method AIM and SCLoss.
We substantiated the effectiveness of our methods by providing extensive empirical studies and theoretical evidence in the Appendix.
These empirical studies include pixel- and slot-space analyses on tasks such as the durability test and property prediction.
Though our work remains several questions detailed in the Appendix and represents just one step on a long journey of OCL, we firmly believe that our work is a foundational piece in the field of interpretable OCL and propel the ongoing effort to equip machines with human-like comprehension abilities.




\newpage
\appendix
\onecolumn

\section{Limitations and Future Works}
\label{sec:limitations}

\noindent\textbf{Advanced object manipulation and real-world datasets.}
In this research, we leverage image augmentation techniques to create pseudo labels for object manipulation at the image level, effectively addressing the absence of object-level ground truths.
Consequently, we gain control over objects within the scope of augmentation-related properties, including attributes such as color, position, and size.
Nonetheless, relying solely on image augmentation to generate supervision signals has its inherent limitations. 
These limitations encompass factors like the diversity of target properties and the extent of manipulation that can be effectively covered.
Exploring avenues for incorporating more informative open-source datasets, such as those used for image captions, holds promise for manipulating object representations more broadly.
We will elaborate this potential in Section \ref{subsec:material_and_shape}.

\noindent\textbf{Position insensitive representation.}
In our research, it is observed that the Slot Attention algorithm generates slots exhibiting sensitivity to the positioning of objects.
Notably, this phenomenon persists even when we exclude the soft positional encoding in the visual encoder.
To achieve a more interpretable object representation, exploring the generation of well-balanced slots across various properties, rather than solely focusing on position, shows potential for future work.

\noindent\textbf{State-aware slot manipulation.} 
In our slot manipulation process, the Property Encoder encodes each property without taking into account the current state of the manipulated slot. 
For instance, when modifying the size of an object, the PropertyEncoder produces a property vector, $\texttt{inst\_vec}$,  irrespective of the color or position of the target object.
By incorporating the current state of the target slot into the slot manipulation process, the precision and complexity of the algorithm could be potentially enhanced. 
\vspace{-0.5em}
\section{Preliminary: Spatial Binding in Slot Attention}
\vspace{-1em}
\begin{algorithm}[H]
	\caption{
 Spatial binding in slot attention algorithm in pseudo-code format. 
 The input image is encoded into a set of $N$ vectors of dimension $D_{input}$ which is mapped to a set of $K$ vectors with dimension  $D_{slot}$. 
 Slots are initialized from a Gaussian distribution with learned parameters $\mu, \sigma \in \mathbb{R}^{D_{slot}}$. 
 The number of iterations is set to $T = 3$.} 
 
	\begin{algorithmic}[1]
        \Function{\text{SpatialBinding}}{$\texttt{img} \in \mathbb{R}^{H \times W \times 3}$}
            \State $\texttt{inputs} = \texttt{Encoder(img)}$
            \State $\texttt{inputs} = \texttt{LayerNorm(inputs)}$
            \For{$t = 0 \ldots T$}
                \State $\texttt{slots\_prev} = \texttt{slots}$
                \State $\texttt{slots} = \texttt{LayerNorm(slots)}$
                \State $\texttt{attn} = \texttt{Softmax}(\frac{1}{\sqrt{D_{slot}}}k(\texttt{inputs})\cdot q(\texttt{slots})^T, \texttt{axis=`slots'})$
                \State $\texttt{updates} = \texttt{WeightedMean(weights=attn+}\epsilon \texttt{, values=} v\texttt{(inputs))}$
                \State $\texttt{slots} = \texttt{GRU(state=slots\_prev, inputs=updates)}$
                \State $\texttt{slots} = \texttt{slots} + \texttt{MLP(LayerNorm(slots))}$
            \EndFor
            \State \textbf{return} \texttt{slots}
        \EndFunction
    \end{algorithmic} 
\label{alg:slota}
\end{algorithm} 


The core mechanism of the slot attention, the spatial binding, is described in Alg. \ref{alg:slota}.
Given an input image \texttt{img} $\in \mathbb{R}^{H \times W \times 3}$, convolutional neural networks (CNNs) encoder generates a visual feature map \texttt{input} $\in \mathbb{R} ^ {N \times D_{enc}}$, where $H$, $W$, $N$, and $D_{\text{enc}}$ are the height and width of the input image, the number of pixels in the input image ($= HW$), and the channel of the visual feature map. 
The slot attention module takes \texttt{slots} and \texttt{inputs}, and projects them to dimension $D_{slot}$ through linear transformations $k$, $q$, and $v$.
Dot-product attention is applied to generate an attention map, \texttt{attn}, with query-wise normalized coefficients, enabling slots to compete for the most relevant pixels of the visual feature map.
The attention map coefficients weight the projected visual feature map to produce updated slots, \texttt{updates}.
With the iterative mechanism of the slot attention module, the slots can gradually refine their representations.


\section{Implementation and experimental details}

\subsection{Training}
We use a single V100 GPU with 16GB of RAM with 1000 epochs and a batch size of 64.
The training takes approximately 65 hours (wall-clock time) using 12GB of RAM for the CLEVR6 dataset, and 22 hours using 9GB of RAM for the Tetrominoes dataset, both with 16-bit precision.

\subsection{Image Augmentation}
Upon receiving an input image $img_{input}$, we produce four outputs: a reference image, denoted as $img_{ref}$, an augmented image, represented as $img_{aug}$, and the transformation instructions between them, indicated as $insts_{ref2aug}$ and $insts_{aug2ref}$.

In the data augmentation process, three pivotal variables are defined. The first is the template size $\mathcal{T}$, employed for the initial cropping of $img_{input}$ prior to the application of transformation (240 for CLEVR6 and 80 for Tetrominoes).
Next, the crop size $\mathcal{C}$ is used to crop the transformed image before resizing it to $\mathcal{M}$ (192 for CLEVR6 and 64 for Tetrominoes). 
This two-stage cropping procedure mitigates the zero-padding that results from transformations. 
Lastly, the image size $\mathcal{M}$ denotes the final image size post data augmentation (128 for CLEVR6 and 64 for Tetrominoes).

In the training phase, $img_{ref}$ is obtained by applying a center-crop operation on $img_{input}$ using $\mathcal{C}$ and then resizing it to $\mathcal{M}$. 
The generation of $img_{aug}$ is more complex, entailing the application of a random transformation from a set of three potential transformations. 
Initially, $img_{input}$ is cropped using $\mathcal{T}$, and the transformation process is implemented. 
Following this, the transformed image is cropped by $\mathcal{C}$ and then resized to $\mathcal{M}$, yielding $img_{aug}$.
The detailed description for each transformation is as follows:

\noindent \textbf{Translating.}
We set a maximum translation value $d_{max} = \frac{\mathcal{T} - \mathcal{C}}{2}$.
A value is randomly chosen within the range of $(-d_{max}, d_{max})$ for translation along the $x$-axis ($d_x$) and the $y$-axis ($d_y$) respectively.


\noindent \textbf{Scaling.}
The maximum and minimum scaling factors, $s_{max}$ and $s_{min}$, are computed by $\frac{\mathcal{T}}{\mathcal{C}}$ and $\frac{\mathcal{C}}{\mathcal{T}}$, respectively.
A float value $s$, serving as a scaling factor, is then randomly sampled from within the range of $(s_{max}, s_{min})$.
One thing to note is that calculating the transformation instructions is not straightforward due to the potential translation of objects during scaling.
Thus, to calibrate the instructions, we infer translation values from the predicted object positions before scaling.
The position prediction is calculated as the weighted mean on the attention maps between the visual encodings and slots. 
With this position prediction, we add the translation term into the scaling process so that the model should perform both object-level scaling and translating: $\vec{d} = (s - 1) (\vec{p} - \vec{c})$, where $\vec{d}$ represents the vector of the translation value, $\vec{p}$ refers to the vector of the predicted object position, and $\vec{c}$ is the vector corresponding to the position of image center.


\noindent \textbf{{Color shifting.}}
In this study, we employ the HSL (hue, saturation, and lightness) color space for effective object color manipulation. 
The input image, initially in RGB space, is converted to HSL space.
We adjust the hue by rotating it using randomly sampled angles that span the entire hue space. 
For saturation, we apply a scaling factor, determined by the exponential of a value randomly drawn from (-1, 1), a hyper parameter. 
Our primary focus lies on the internal color of objects, leaving lightness untouched. 
Nonetheless, adjustments to lightness can be made if necessary.

\noindent \textbf{Instruction.}
Each transformation instruction is a list of 6 values: one scaling factor $(\Lambda \text{scale})$, two translation parameters $(\Delta x, \Delta y)$, and three color shifting parameters in HSL $(\Delta \text{hue}, \Lambda \text{saturation}, \Lambda \text{lightness})$ where $\Lambda$ and $\Delta$ means the multiplicative and additive factor for the corresponding values, respectively.
The identity instruction, $insts_{identity}$, contains the base values for each transformation. 
Thus, $insts_{identity}$ has $1$ for scaling, $(0, 0)$ for translation, and $(0, 1, 1)$ for color shifting.
For the inverse instruction , $insts_{aug2ref}$ has the values of $-insts_{ref2aug}$ for additive factors, and $\frac{1}{insts_{ref2aug}}$ for multiplicative factors.

\subsection{Model}
Basically, our model framework is built on Slot Attention \citep{locatello2020object}, thereby the encoder, decoder, and slot attention module are the same as that of Slot Attention except for the inclusion of the Attention Refining Kernel (ARK) from SLASH \citep{kim2023shepherding}.
For Tetorminoes and CLEVR, we employ a 4-layer CNN encoder and a 6-layer Spatial Broadcast (SB) decoder \citep{watters2019spatial} with a hidden dimension of 64. 
Within the slot attention module, we set the slot dimension to 64, perform the binding process for 3 iterations, and use a kernel size of 5 for the ARK.
Please refer to the original papers \citep{kim2023shepherding,locatello2020object} for additional details for Slot Attention.

For CLEVRTEX6 and PTR datasets which include more complicated objects, we adopt larger models with a slot dimension, $D_{slot}$, of 256.
As encoders, we use 1) Resnet34 \citep{he2016deep} following \citep{elsayed2022savi++, biza2023invariant} and 2) ViT-base \citep{dosovitskiy2020image}, with the patch size of 8, pretrained via MAE \citep{he2022masked}
As decoders, we use an increased size of SB decoder consisting of 8-layer CNNs with a hidden dimension of 128, and a Transformer-based decoder proposed in SRT \citep{sajjadi2022scene}.
The original SRT decoder is designed to operate at the image level, and the following research OSRT \citep{sajjadi2022object} introduce a modification to decode slots simultaneously. 
In this paper, we slightly modified it to decode each slot independently following the spatial broadcast decoder.
This selection is made to demonstrate that our proposed method is not limited to CNN-based spatial broadcast decoders used in Slot Attention but can robustly operate within transformer-based decoders as well, given the appropriate conditions for independence.
The results of using large models are described in Sec. \ref{sec:further_experiments}.

In Alg. 2 of the main paper, the Property Encoder ($\texttt{PropertyEncoder}$) takes as input the values that correspond to specific properties. 
Accordingly, the input size for the property encoder is 1 for scaling, 2 for translation, and 3 for color shifting. 
Each property is encoded via Property Encoder, a 3-layer MLP with ReLU activation functions, resulting in a $\texttt{inst\_vec}$, a vector of dimension $D_{slot}$.


\section{From the Image-level Training to Object-level Inference}

To begin with, we would like to highlight our unique approach to the training procedure.
While our training incorporates manipulations at the \textit{image-level}, it can be perceived as training the model at the individual \textit{object-level}.
In this section, we discuss on how this transition is achieved without the need for an additional tuning process, and present empirical results that support our claim.

As we discussed shortly in Sec. 3.1. in the main paper, the success of transitioning from image-level augmentation during training to object-level manipulation during inference can be attributed primarily to the fact that the entire process for each slot, including object discovery and decoding, exclusively influences the reconstruction of its respective \textit{object}.
To substantiate our claim, a mathematical proof is provided below to show how an image-level reconstruction loss can be disentangled into object-level reconstruction losses.

\begin{equation}
\mathcal{L}_{\text{recon}} = \| \hat{\mathcal{I}} - \mathcal{I} \|_{2}^{2}
\label{eq:line1}
\end{equation}

\begin{equation}
= \| \sum_{k=1}^{\mathcal{K}} (\hat{\mathcal{I}}_{k}^{rgb} \odot \hat{\mathcal{I}}_{k}^{\alpha}) - \mathcal{I} \|_{2}^{2}
\label{eq:line2}
\end{equation}

\begin{equation}
= \| \sum_{k=1}^{\mathcal{K}} (\hat{\mathcal{I}}_{k}^{rgb} \odot \hat{\mathcal{I}}_{k}^{\alpha}) - \sum_{k=1}^{\mathcal{K}} (\mathcal{I} \odot \hat{\mathcal{I}}_{k}^{\alpha}) \|_{2}^{2}
\label{eq:line3}
\end{equation}

\begin{equation}
= \| \sum_{k=1}^{\mathcal{K}} (\hat{\mathcal{I}}_{k}^{rgb} \odot \hat{\mathcal{I}}_{k}^{\alpha} - \mathcal{I} \odot \hat{\mathcal{I}}_{k}^{\alpha}) \|_{2}^{2}
\label{eq:line4}
\end{equation}

\begin{equation}
\approx \| \sum_{k=1}^{\mathcal{K}} (\hat{\mathcal{O}}_{k} - \mathcal{O}_{k}) \|_{2}^{2},
\label{eq:line5}
\end{equation}

\begin{equation}
= \sum_{k=1}^{\mathcal{K}} \| (\hat{\mathcal{O}}_{k} - \mathcal{O}_{k}) \|_{2}^{2}
+ \sum_{\substack{i,j=1 \\ i \neq j}}^{\mathcal{K}}  (\hat{\mathcal{O}}_{i} \cdot \hat{\mathcal{O}}_{j} - 2 \ \hat{\mathcal{O}}_{i} \cdot \mathcal{O}_{j} + \mathcal{O}_{i} \cdot \mathcal{O}_{j})
\label{eq:line6}
\end{equation}

\begin{equation}
\approx \sum_{k=1}^{\mathcal{K}} \| (\hat{\mathcal{O}}_{k} - \mathcal{O}_{k}) \|_{2}^{2},
\label{eq:line7}
\end{equation}

where $\mathcal{K}$ is the number of slots, $\hat{\mathcal{I}} \in \mathbb{R}^{H \times W \times 3}$ represents the reconstructed image, and $\mathcal{I} \in \mathbb{R}^{H \times W \times 3}$ represents the input image. 
$\hat{\mathcal{I}}_{k}^{rgb} \in \mathbb{R}^{H \times W \times 3}$ and $\hat{\mathcal{I}}_{k}^{\alpha} \in \mathbb{R}^{H \times W \times 1}$ are the reconstruction results generated by the decoder using the k-th slot as input: an RGB and an alpha map (or an attention mask), respectively.
$\hat{\mathcal{O}}_{k} \in \mathbb{R}^{H \times W \times 3}$ is the predicted image for the specific object that is bounded with the k-th slot, while $\mathcal{O}_{k} \in \mathbb{R}^{H \times W \times 3}$ is the corresponding ground-truth object image.

From Eq. (\ref{eq:line1}) to Eq. (\ref{eq:line2}), we follow the decoding process of Slot Attention \citep{locatello2020object}.
In particular, each k-th slot is decoded independently, resulting in the reconstructed RGB image $\hat{\mathcal{I}}_{k}^{rgb}$ and the reconstructed alpha map $\hat{\mathcal{I}}_{k}^{\alpha}$. 
The final reconstruction image $\hat{\mathcal{I}}$ is generated by aggregating $\hat{\mathcal{I}}_{k}^{rgb}$ using a pixel-level weighted average, where the weights are determined by $\hat{\mathcal{I}}_{k}^{\alpha}$.
It is crucial to recognize that $\hat{\mathcal{I}}_{k}^{\alpha}$ serves as an attention mask, as elaborated below:
\begin{equation}
\sum_{k=1}^{\mathcal{K}} \hat{\mathcal{I}_k^{\alpha}}(x, y) = 1 
\phantom{00} \text{for all} \phantom{0} x, y,
\label{eq:calibration}
\end{equation}

where $\hat{\mathcal{I}}_{k}^{\alpha}(x, y)$ is a value for the position $(x, y)$.
This characteristic plays a pivotal role in our approach, facilitating the transition from Eq. (\ref{eq:line2}) to Eq. (\ref{eq:line3}).
In this transformation, the input image $\mathcal{I}$ is effectively weighted by the set of $\mathcal{K}$ alpha maps, denoted as $\hat{\mathcal{I}}_{k}^{\alpha}$, where $k$ spans from 1 to $\mathcal{K}$. 
Then, as both the first and second terms in Eq. (\ref{eq:line3}) involve the same sigma operations, we can simplify the expression by combining the individual subtraction operations into a single sigma operation (Eq. (\ref{eq:line4})).

\begin{figure}
    \centering
    \includegraphics[width=0.9\linewidth]{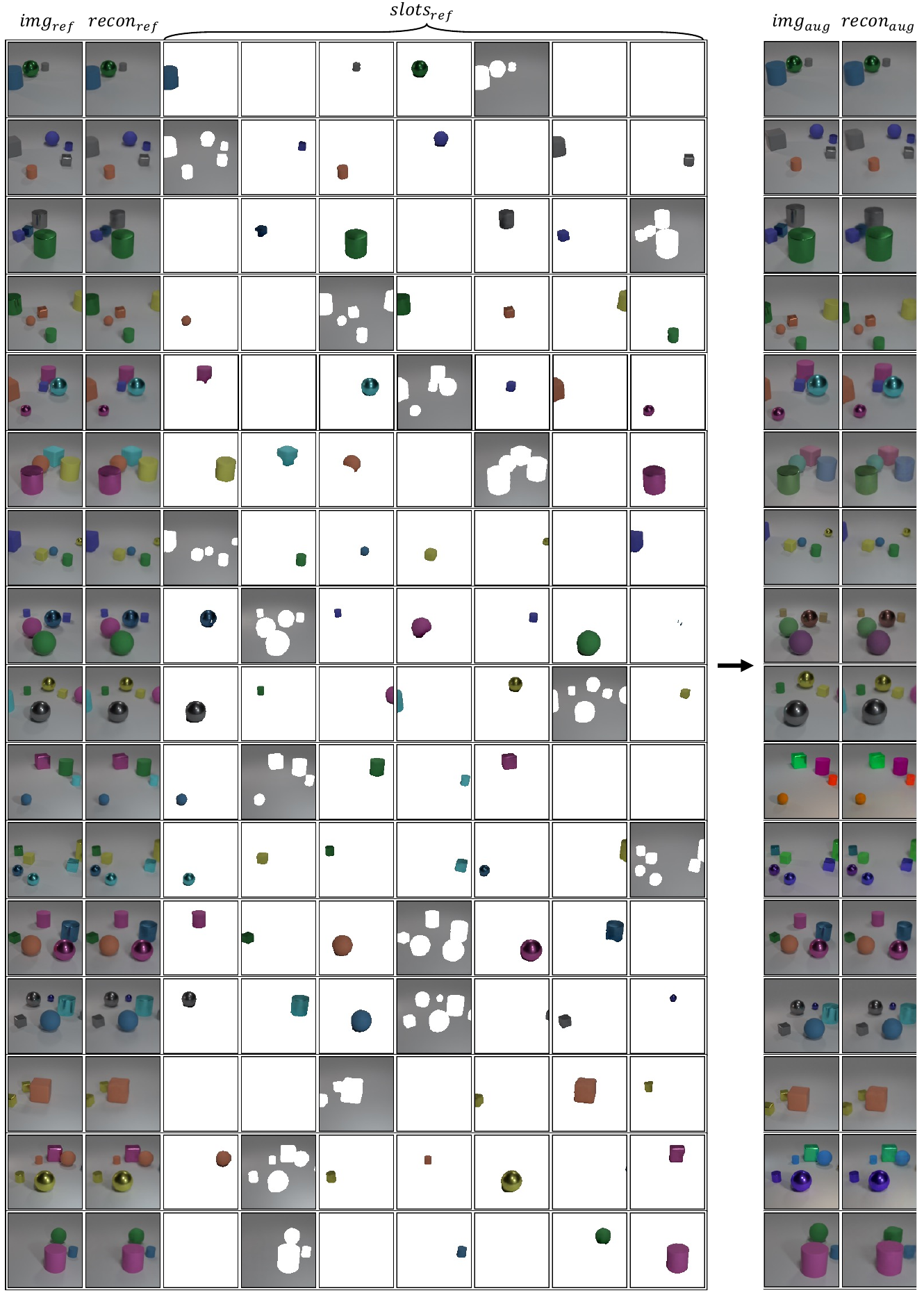}
    \vspace{-1em}
    \caption{ 
    \textbf{Training results of our method.}
    The leftmost column is the reference images, $img_{ref}$. 
    The second leftmost column is the reconstruction of the reference images, $recon_{ref}$.
    The middle columns show the object discovery results where each column corresponds to a single slot in $slots_{ref}$.
    The second rightmost column is the augmented images, $img_{aug}$.
    The rightmost column is the reconstruction of the augmented images, $recon_{aug}$. 
    }
    \label{fig:slotaug}
\end{figure}

Subsequently, we approximate Eq. (\ref{eq:line4}) as Eq. (\ref{eq:line5}) to get an object-level disentangled version of the reconstruction loss.
Here we assume that both $\hat{\mathcal{O}}_{k}$ and $\mathcal{O}_{k}$ only consist of a specific region of interest within the input image.
This region corresponds to the target object which is bound to the $k$-th slot, while the remaining areas are masked out and assigned a value of zero.
We can make this assumption based on the successful performance of the previous object-centric learning model, SLASH \citep{kim2023shepherding}. 
SLASH has demonstrated effective capabilities in focusing on and capturing specific objects of interest within an image, by introducing the Attention Refining Kernel (ARK).
By incorporating ARK into our model, we confidently assume that $\hat{\mathcal{O}}_{k}$ and $\mathcal{O}_{k}$ primarily represent the target object while masking out other irrelevant parts as zero as shown in Fig. \ref{fig:slotaug}.

\begin{figure}
    \centering
    \includegraphics[width=\linewidth]{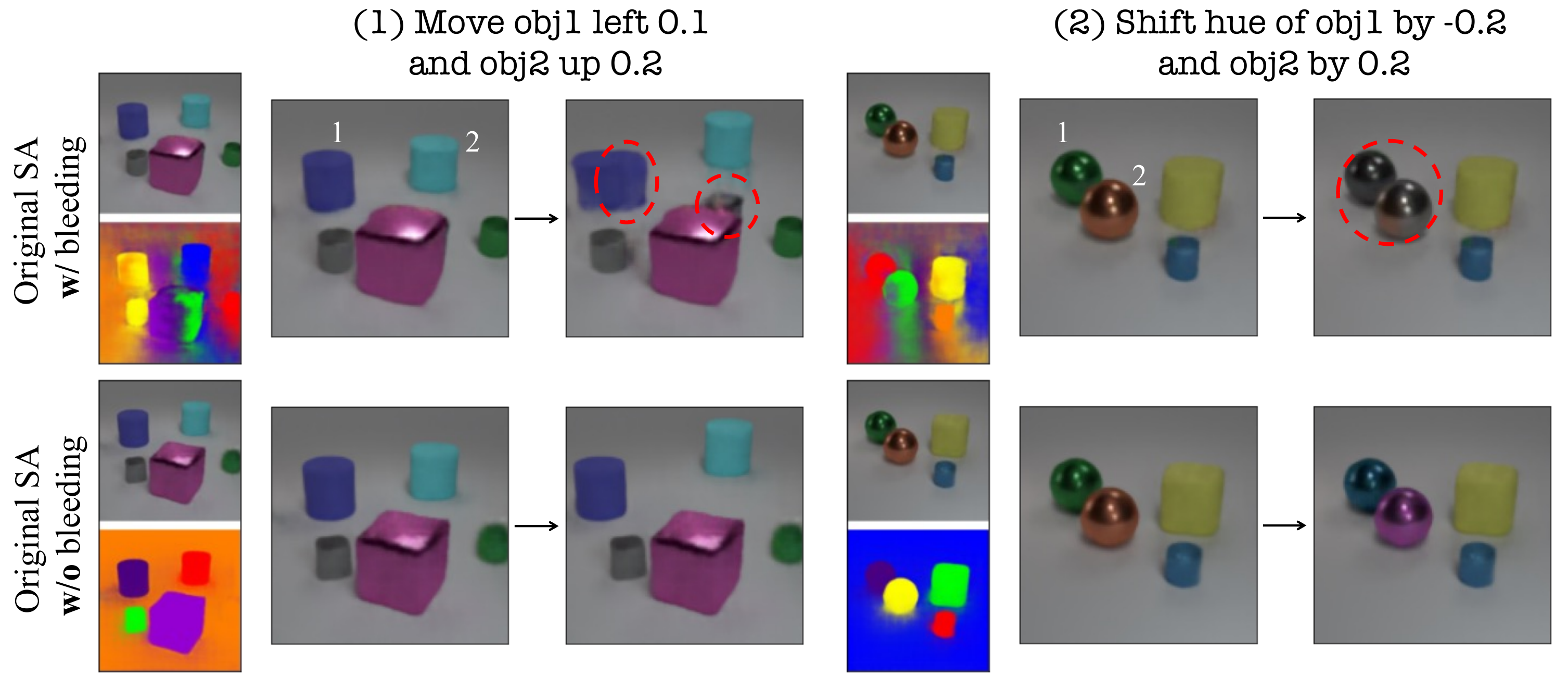}
    \vspace{-1em}
    \caption{ 
    \textbf{Visualization of object manipulation results affected by the bleeding problem with the original Slot Attention.}
    The first row demonstrates the cases where bleeding problem emerges, while the second row shows the cases where the object discovery is done successfully.
    }
    \label{fig:bleeding}
\end{figure}

Here, we would like to note that ARK is an optional component in our method, not a necessity.
The use of ARK is not intended to enhance object discovery performance in a single training session; rather, it is employed to ensure consistent results across multiple experiments.
If our proposed training scenario arises where bleeding issues do not occur in the original SA, it can be achieved without the need for ARK.
To substantiate this claim, we present qualitative results in Fig. \ref{fig:bleeding}, where we train SlotAug with the original SA (without ARK). 
One can easily catch that the object manipulation fails in the case of bleeding problem.
Specifically, the analysis for the failure case in bleeding problem is as follows:
1) Obviously, if the attention map corresponding to the target object encompasses other objects, it becomes impossible to exclusively manipulate solely the target object, leading to unexpected artifacts in other objects.
2) Whenever tinting instructions are applied, objects become gray and we attribute this to the backgrounds -- having a gray color -- intervening with the target objects during training.

Eq. (\ref{eq:line5}) can be broken down into two separate summations.
The first one is our target term that is the sum of object-level MSE losses, and the second term is the residual term.
Lastly, the transition from Eq. (\ref{eq:line6}) to Eq. (\ref{eq:line7}) constitutes a significant simplification in the representation of the loss function.
This is a valid transformation under the assumption follows:

\begin{equation}
\hat{\mathcal{O}}_{i} \cdot \hat{\mathcal{O}}_{j}
= \hat{\mathcal{O}}_{i} \cdot \mathcal{O}_{j}
= \mathcal{O}_{i} \cdot \mathcal{O}_{j}
= 0
\phantom{00} \text{if} \phantom{0} i \neq j.
\label{eq:orthogonal}
\end{equation}

This assumption postulates that the inner product of different object images, whether they are predicted or ground-truth, is always zero. 
We assert that this assumption is justifiable, much like the previous one, given the promising results obtained in our object discovery experiments.
The loss computation is thus decomposed into individual components for each slot, which lends itself to an interpretation of object-level loss.

The conversion from image-level MSE loss to a sum of individual object-level MSE losses provides a new perspective on our training method.
Despite the use of image-level manipulations, the underlying core of the training process inherently engages with object-level representations.
This demonstrates how a simple methodological addition, incorporating image augmentation into the training process, can lead to considerable gains in the model's capacity for user-intention-based object manipulation.

\begin{figure}
    \centering
    \includegraphics[width=0.9\linewidth]{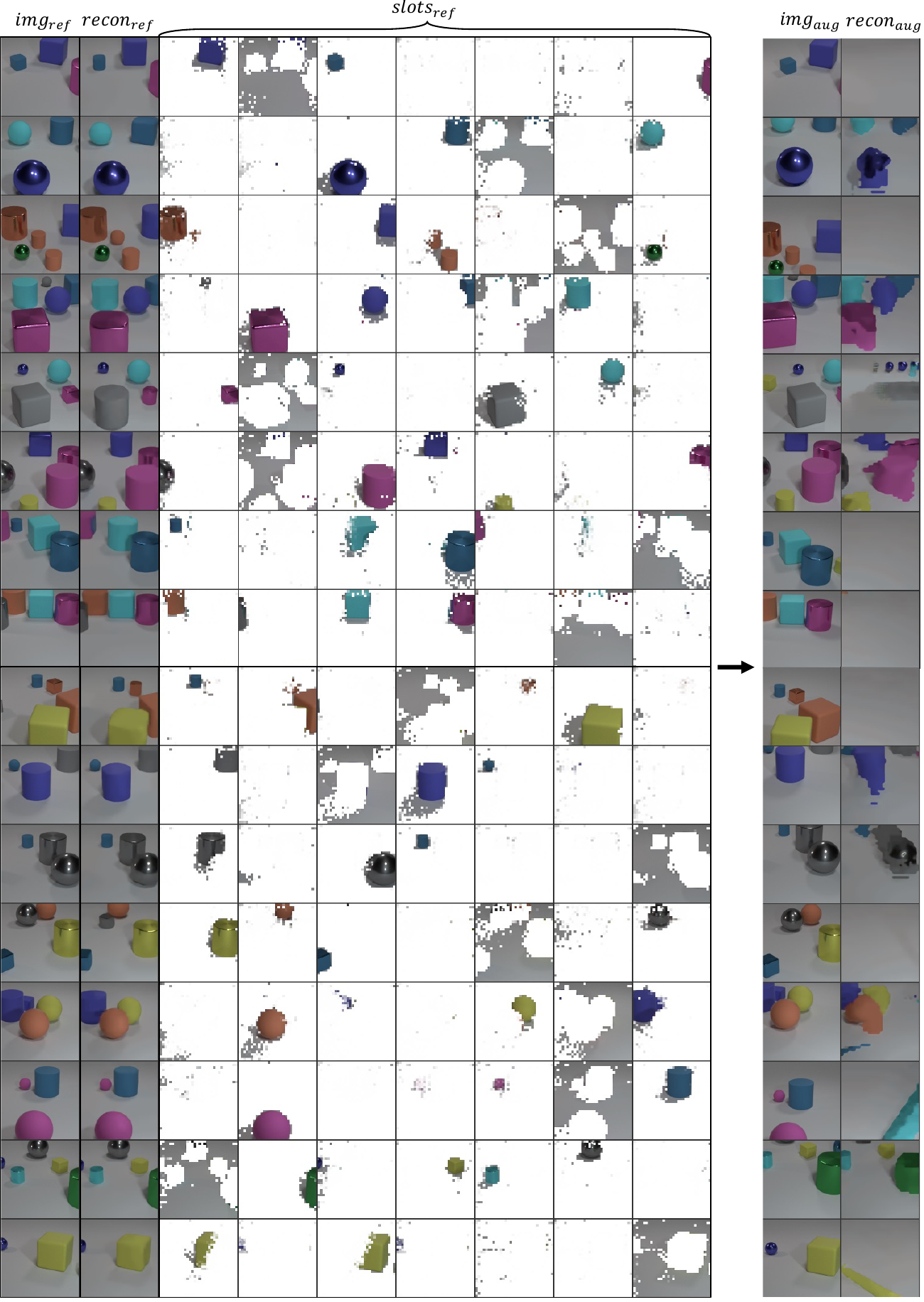}
    \vspace{-1em}
    \caption{ 
    \textbf{Training results of SLATE \citep{singh2021illiterate} for slot manipulation.}
    The leftmost column is the reference images, $img_{ref}$; 
    The second leftmost column is the reconstruction of the reference images, $recon_{ref}$.
   The middle columns show the object discovery results where each column corresponds to a single slot in $slots_{ref}$.
    The second rightmost column is the augmented images, $img_{aug}$.
    The rightmost column is the reconstruction of the augmented images, $recon_{aug}$. 
    }
    \label{fig:slate}
\end{figure}

Fig. \ref{fig:slotaug} empirically demonstrates the effectiveness of our model, leveraging Slot Attention for controllability over slots. 
Conversely, it was noted that the well-known alternative framework for object-centric learning, SLATE \citep{singh2021illiterate}, employing image tokenization from Discrete VAE (dVAE) \citep{im2017denoising} and Transformer-based auto-regressive decoding \citep{vaswani2017attention}, struggled with the manipulation of slots, as illustrated in Fig. \ref{fig:slate}. 
The same slot manipulation strategy via Property Encoder was used for comparison. 
Other training environments are just the same as the official paper \citep{singh2021illiterate} except for the addition of the training loss for the reconstruction of the augmented images.





\section{Further experimental results}
\label{sec:further_experiments}

\begin{figure}
    \centering
    \includegraphics[width=\linewidth]{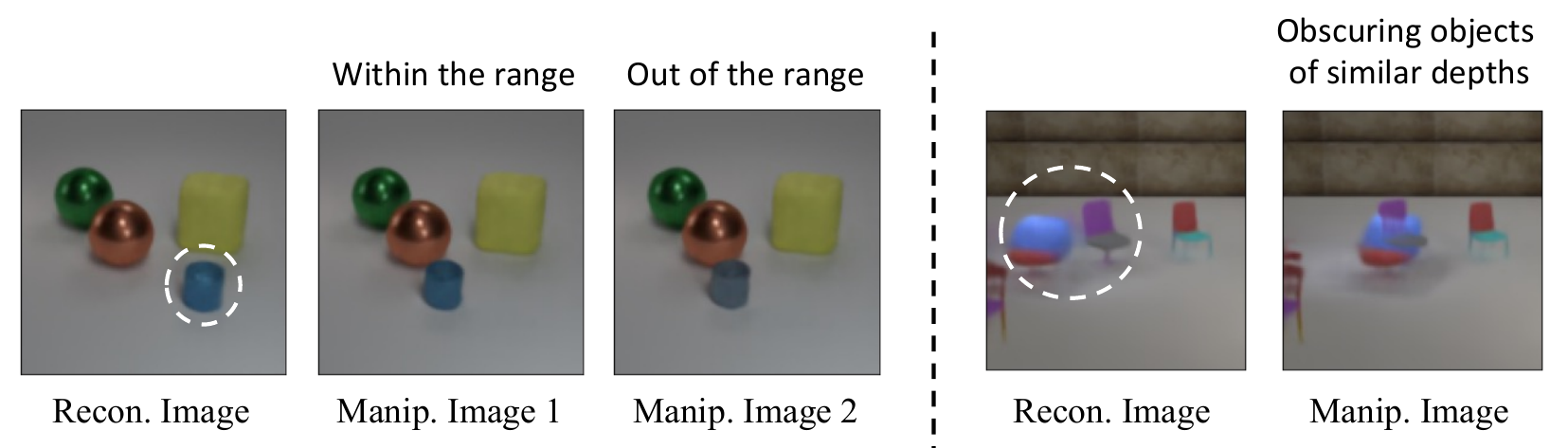}
    \vspace{-1em}
    \caption{ 
    \textbf{Visualization of failure cases.}
    On the left, the target object loses its original color when translated beyond the range defined in the training settings. 
    On the right, unnatural overlapping between objects occurs when objects have similar depths.
    }
    \label{fig:failure}
\end{figure}

\begin{figure}
    \centering
    \includegraphics[width=0.9\linewidth]{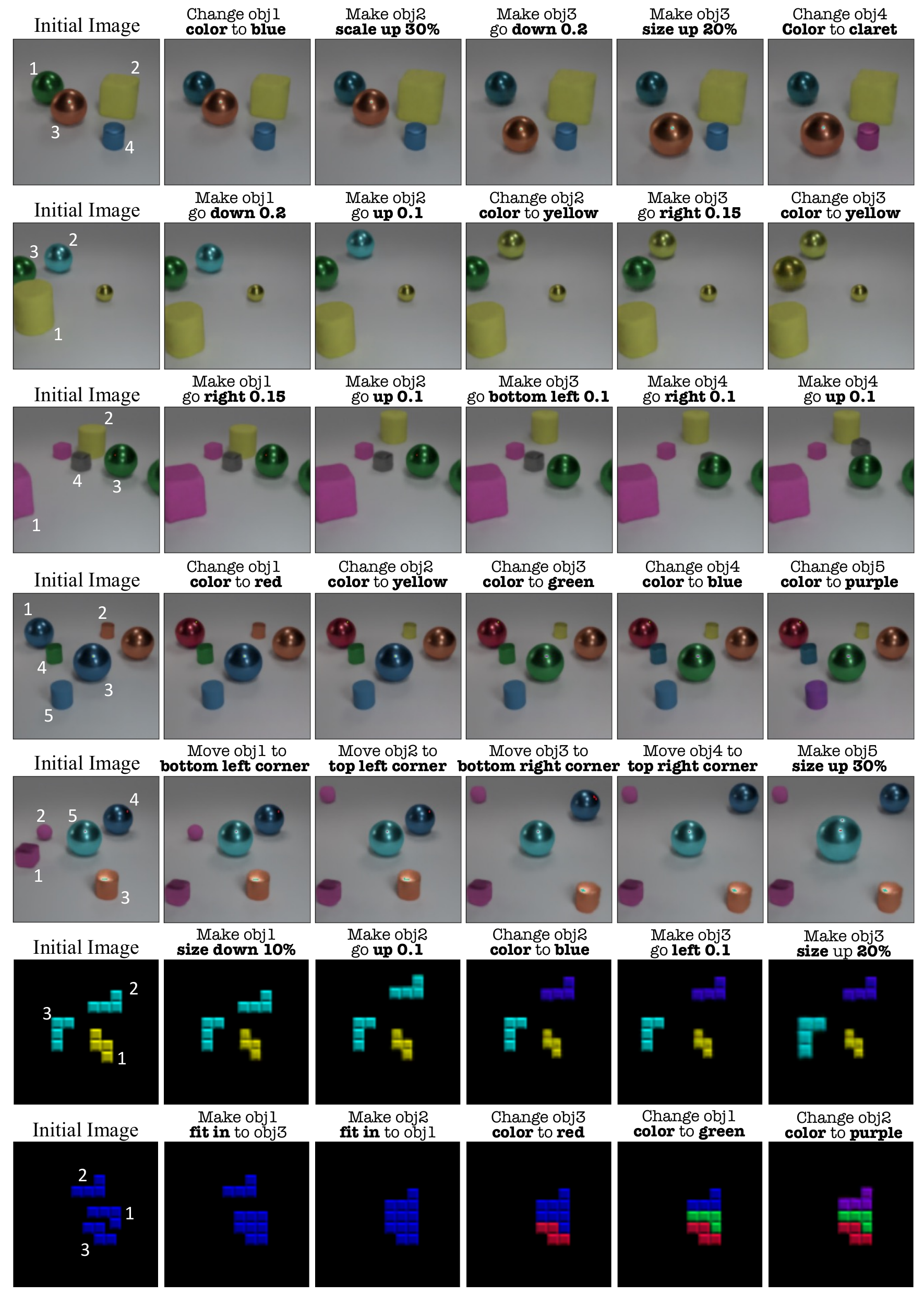}
    \vspace{-1em}
    \caption{ 
    \textbf{Visualization of object manipulation in CLEVR6 and Tetrominoes dataset.}
    The leftmost column features the initial images, serving as the starting point for the manipulation process. The subsequent columns depict the results of object-level manipulation, following the instructions represented as text above the images.
    }
    \label{fig:obj_manip}
\end{figure}

\begin{figure}
    \centering
    \includegraphics[width=\linewidth]{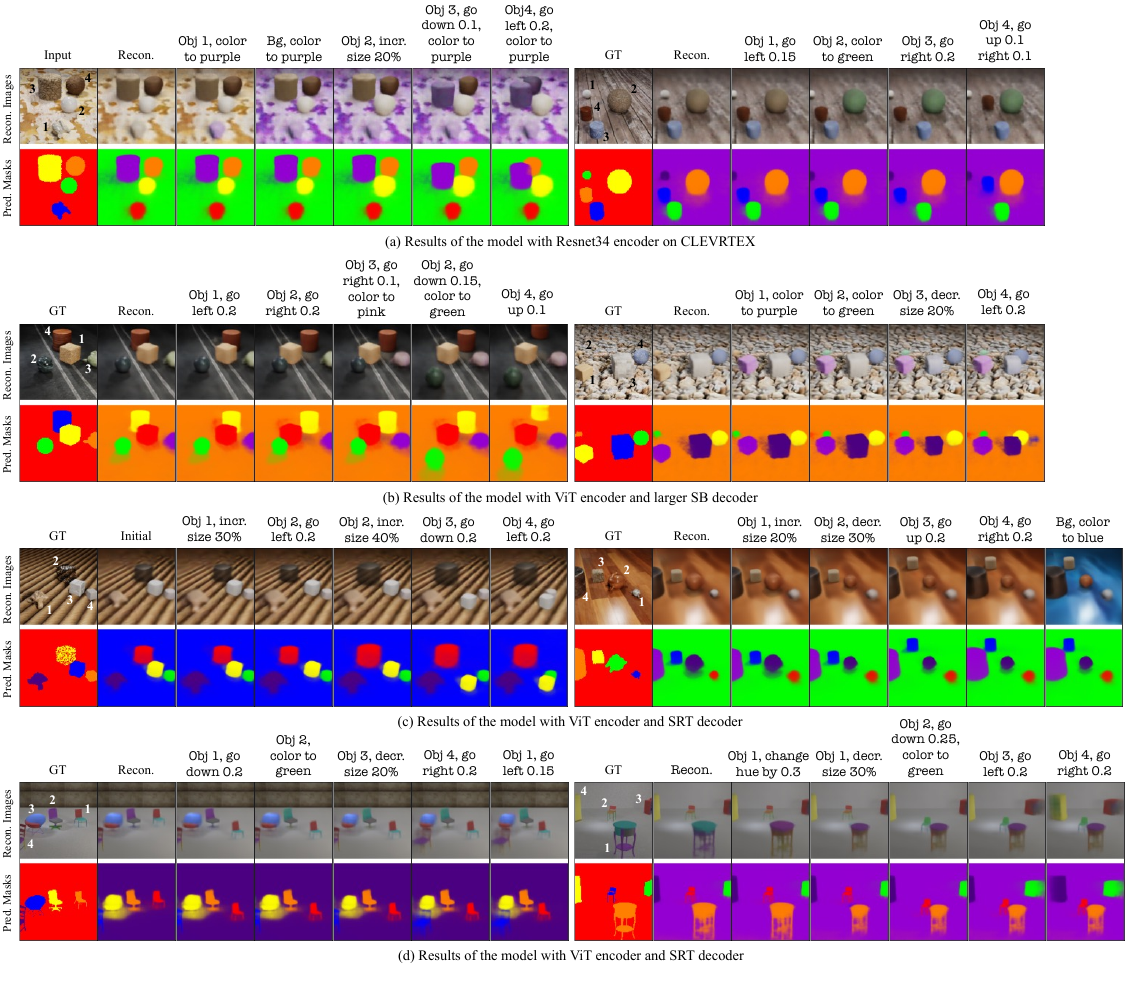}
    \vspace{-1em}
    \caption{ 
    \textbf{Visualization of object manipulation in CLEVRTEX and PTR with larger encoder and decoder.}
    The first three rows are for CLEVRTEX and the last is for PTR dataset. Regarding the encoder, we initialize ResNet34 randomly, while ViT is pre-trained using MAE \citep{he2022masked} on the target datasets. For the decoder, we employ both the Spatial Broadcast (SB) decoder and the SRT decoder \citep{sajjadi2022scene}. In case (b), we enhance the size of the SB decoder with a hidden dimension of 128 and a depth of 8. Additionally, for SRT, we adopt a slot-wise decoding strategy akin to the SB decoder.
    }
    \label{fig:obj_manip_complex}
\end{figure}

In this section, we present more qualitative results including failure cases (Fig. \ref{fig:failure}), and object-level manipulation using various backbone encoders and decoders (Fig. \ref{fig:obj_manip} and \ref{fig:obj_manip_complex}).
Furthermore, we showcase several additional experimental results below.

\subsection{Fully Supervised Training on Materials and Shapes}
\label{subsec:material_and_shape}

To explore the capabilities of our method when provided with human-annotated labels, we demonstrate object manipulation examples related to materials and shapes.
We utilize the CLEVR render \footnote{\label{footnote:clevr-renderer}https://github.com/facebookresearch/clevr-dataset-gen} to generate datasets having ground truth in terms of property modification.
Using datasets containing precise annotations for the target properties (materials and shapes), we can explicitly train a model through object-level supervision.

\begin{figure}
    \centering
    \includegraphics[width=\linewidth]{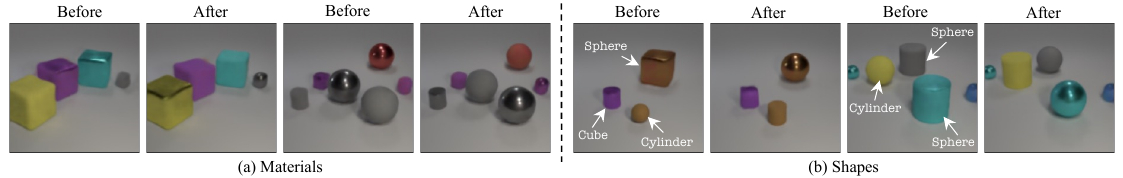}
    \vspace{-1em}
    \caption{ 
    \textbf{Results of manipulating materials and shapes of objects.} For this experiment, the training datasets are crafted using the CLEVR renderer, wherein we modify target properties such as materials and shapes while keeping other properties unchanged.
    }
    \label{fig:material_shape}
\end{figure}

As shown in Fig. \ref{fig:material_shape}, the model can effectively acquire knowledge of extrinsic properties, such as material and shape, when provided with appropriate supervision signals.
In future work, one can aim to enhance the proposed training scheme by leveraging more informative datasets, such as those for image captioning \citep{chen2015microsoft, wang2023all}, to train a more human-interactive framework.
This process may entail elaborate data processing since the datasets were not originally designed for the purposes of object manipulation.
However, we firmly believe that pursuing this path holds great promise.

\subsection{Quantitative Evaluation on Object Manipulation}

\begin{table}[]
\begin{center}
\caption{
\textbf{Results of object-level manipulation on the rendered dataset.}
We evaluate the object-level manipulation by assessing metric scores over three sorts of images: reference images (ref.); manipulated images (manip.); and reversion or restored images (rev.).
We use mIoU and ARI for the object discovery task, and MSE for the image generation task.}

\label{tab:quantitative}
\begin{tabular}{lccccccccccc}
\toprule
 & \multicolumn{3}{c}{mIoU} &  & \multicolumn{3}{c}{ARI} &  & \multicolumn{3}{c}{MSE} \\ \cline{2-4} \cline{6-8} \cline{10-12} 
 & ref. & manip. & rev. &  & ref. & manip. & rev. &  & ref. & manip. & rev. \\ \hline
v1 & 89.4 & 69.9 & 68.9 &  & 97.9 & 79.3 & 80.1 &  & 7.3e-4 & 4.6e-3 & 6.3e-3 \\
v2 & 87.9 & 71.4 & 80.0 &  & 96.1 & 79.4 & 90.2 &  & 7.5e-4 & 4.2e-3 & 3.1e-3 \\
v3 & 85.4 & 70.5 & 79.8 &  & 95.3 & 78.8 & 89.8 &  & 7.8e-4 & 3.5e-3 & 2.1e-3 \\
\bottomrule
\end{tabular}
\end{center}
\end{table}

The supplementary quantitative results on object manipulation are shown in Tab. \ref{tab:quantitative}.
Given the absence of an evaluation benchmark dataset, we opt to employ the same CLEVR render as described in Sec. \ref{subsec:material_and_shape}, to generate a set of 1500 triplets. 
These triplets consist of a reference image, instructions for object manipulation, and the resulting manipulated image.
Additionally, it is worth noting that there exists no prior research specifically addressing slot manipulation through direct human-interpretable instructions. 
Consequently, our performance comparisons are restricted to different versions of our model: v1 (the base model with image augmentation only), v2 (image augmentation + AIM), and v3 (image augmentation + AIM + SCLoss).

One can observe that all three models successfully execute object discovery and object manipulation tasks with minimal differences in their performance scores, highlighting the effectiveness of our training approach leveraging image augmentation.
However, in the context of the reversion task, wherein the models are instructed to revert the manipulated objects to their original state, both v2 and v3 outperform v1, demonstrating the effectiveness of the proposed AIM. 
Furthermore, in terms of image editing, v3 surpasses both v1 and v2 by a significant margin, underscoring the effectiveness of the proposed SCLoss.

\subsection{Ablation Study on Loss Weihgts}
We conduct an ablation study on training losses using the v3 model (image augmentation + AIM + SCLoss). 
Tab. \ref{tab:loss_weights} shows the results of training models with three different loss weights for the SCLoss ($w_{cycle}$) while maintaining the $w_{recon}$ as 1.0.
For the balanced training result, considering both image reconstruction and object discovery, we opted for 0.1 due to its balanced performance.

\begin{table}[]
\begin{center}
\caption{
\textbf{Results of ablation studies on weights for the training loss.}
The leftmost column shows the values of the weight for SC-Loss ($w_{cycle}$), while the weight for the reconstruction loss ($w_{recon}$)is set to $1.0$.
The other columns show the training losses when training is finished and the scores of the validation metrics for the object discovery task.
Each row shows the results of using the weight for SC-Loss with $1.0$, $0.1$, and $0.01$, respectively.
}
\label{tab:loss_weights}
\begin{tabular}{lcccccc}
\toprule
\multicolumn{1}{c}{\multirow{2}{*}{$w_{cycle}$}} & \multicolumn{3}{c}{Train} & \multicolumn{2}{c}{Val} \\ \cline{2-6} 
\multicolumn{1}{c}{} & loss\_recon\_ref & loss\_recon\_aug & loss\_cycle & mIoU & ARI \\ \hline
1.0* & 3.5e-4 & 4.9e-4 & 2.2e-6 & 66.5 & 94.1 \\
0.1 & 3.2e-4 & 3.8e-4 & 1.8e-5 & 68.5 & 95.2 \\
0.01 & 3.1e-4 & 4.4e-4 & 6.2e-5 & 68.7 & 95.4 \\
\bottomrule
\end{tabular}
\end{center}
\end{table}

\subsection{Extreme Durability Test}
\begin{figure}
    \centering
    \includegraphics[width=\linewidth]{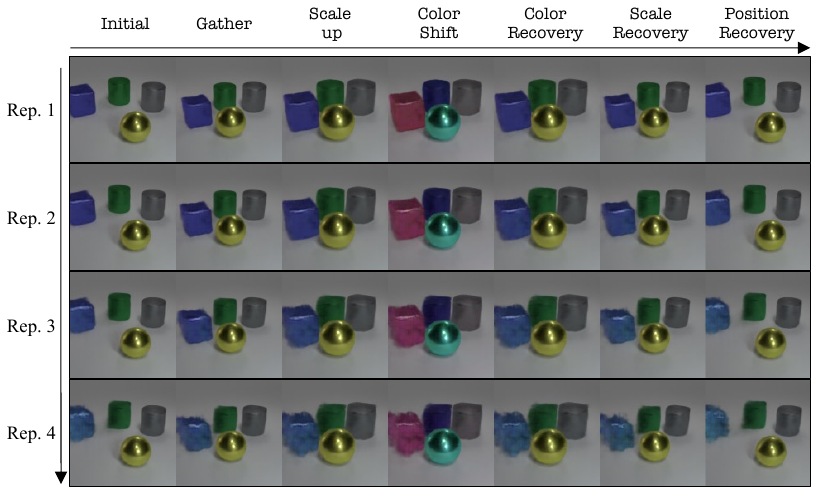}
    \vspace{-1em}
    \caption{ 
    \textbf{The results of the durability test} are depicted in the figure, wherein all objects in the scene are manipulated in accordance with the instructions, provided as text at the top of the figure. 
    The leftmost column presents the initial state of the image. 
    The subsequent three columns comprise three distinct manipulations: translation, scaling, and color shifting. 
    The right three columns feature three recovery manipulations intended to restore the image to its original state. 
    We perform 4 cycles of these round trip processes, leading to a total of 24 manipulations.
    }
    \label{fig:durability1}
\end{figure}

We evaluate our v3 model with two stringent versions of the durability test. 
Fig. \ref{fig:durability1} displays the first extreme durability test, wherein we manipulate all objects within a given scene across a total of 24 manipulation steps. 
The complete manipulation process encompasses four cycles of round-trip manipulations, each cycle comprising three sequential forward manipulations followed by three recovery manipulations. 
Despite the fact that the appearance of each object tends to deviate from its initial state as manipulations accumulate, it is notable that our model demonstrates substantial robustness against multiple rounds of manipulations.

\begin{figure}
    \centering
    \includegraphics[width=\linewidth]{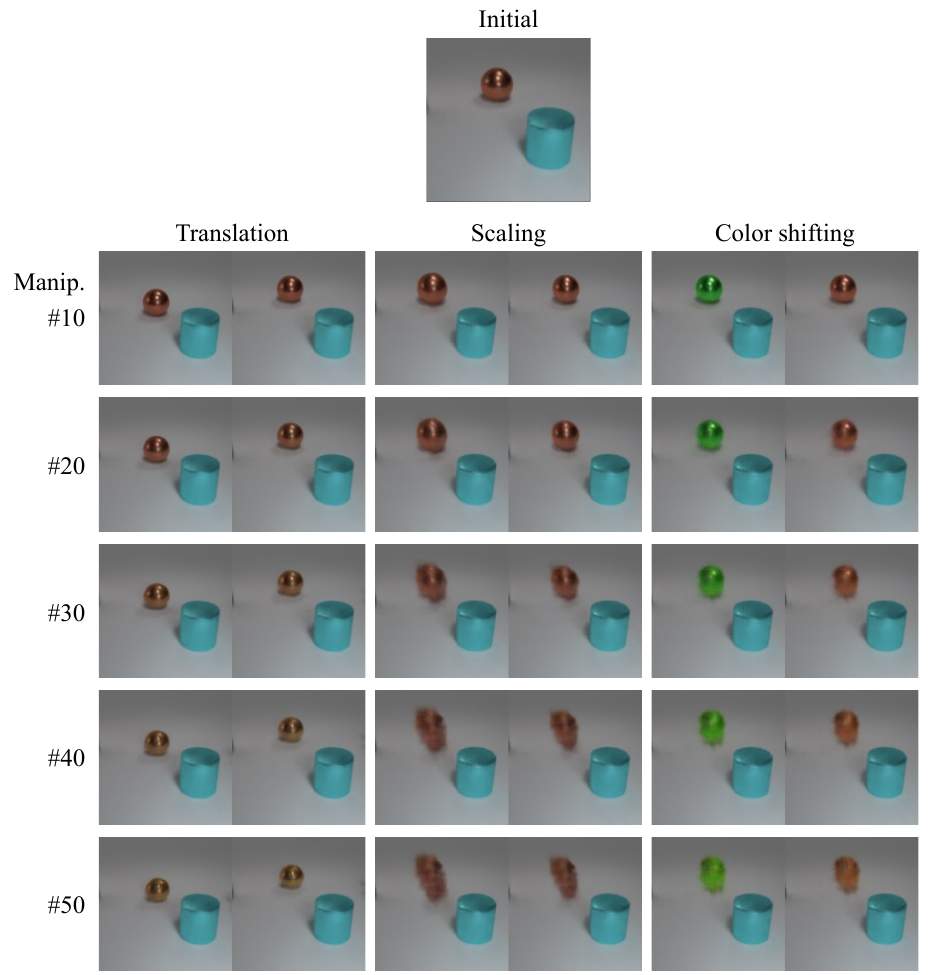}
    \vspace{-1em}
    \caption{
        \textbf{The results of the single-step durability test.}
        The top image represents the initial state prior to any manipulation.
        Each column depicts the results of the single-step durability test with translation, scaling, and color shifting, respectively.
        The left images in each column illustrate the outcome of the manipulation corresponding to the column name, while the right images in each column display the results of the recovery, or inverse, manipulation.
        Each row represents the results after a series of manipulations, with the number of manipulations corresponding to the row number.
    }
    \label{fig:durability2}
\end{figure}

In the second durability test, we manipulate a target object through 50 steps of manipulation, which consists of 25 cycles of a single forward manipulation (translation, scaling, or color shifting) and its corresponding recovery manipulation. 
Fig. \ref{fig:durability2} demonstrates our model's significant endurance against numerous manipulation steps. We observe that our model exhibits greater robustness in translating objects compared to scaling and color shifting. 
However, it is noteworthy that the resilience of our model against both scaling and color shifting is impressive, as it withstands around 20 steps of manipulations without significant distortion in object appearance.
The figure further includes qualitative results that gauge slot divergence. 
It becomes clear that our model's durability improves gradually as it evolves from version v1 to v2 and finally to v3.

\subsection{Toy Application: Object-centric Image Retrieval}
With the acquisition of object-level controllability, we can extend object-centric learning to a newly introduced downstream task, called object-centric image retrieval.
Object-centric image retrieval aims to retrieve an image having an object that is most relevant to a target object that is given by the user's intention.

\begin{figure}
    \centering
    \includegraphics[width=\linewidth]{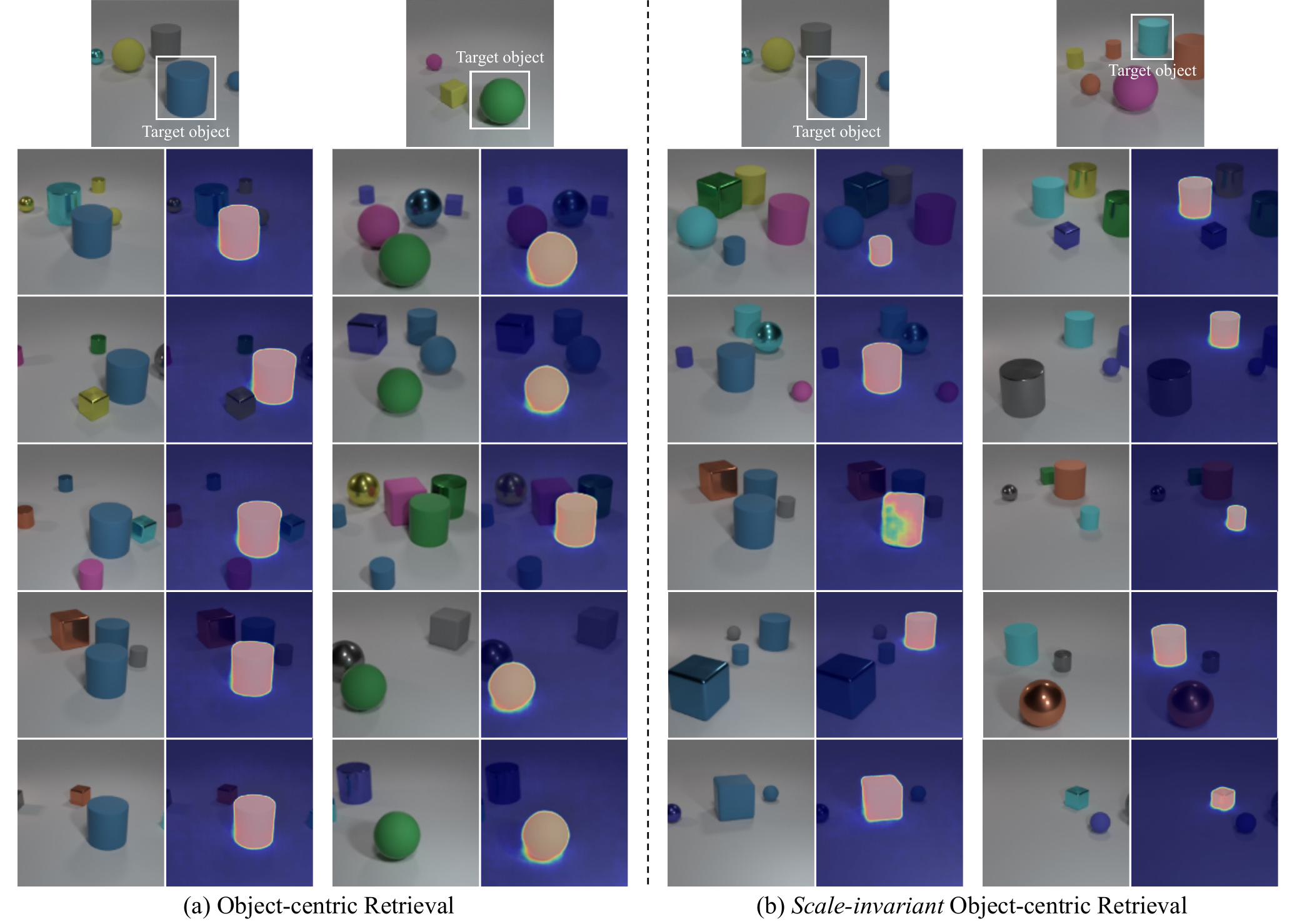}
    \vspace{-1em}
    \caption{ 
        \textbf{Visualization of object-centric image retrieval.}
        The top row displays query images, indicating the target objects to be retrieved. 
        Below each query image, you can find the top 5 retrieval results. 
        Each retrieval result consists of the original image on the left and an attention map on the right. 
        The attention map, associated with the slot corresponding to the target object, emphasizes the specific region within the image.
    }
    \label{fig:retrieval}
\end{figure}

The retrieval process is as follows.
First, we acquire slots for a target object and candidate objects from the corresponding images by conducting object discovery.
Then, to reduce the effect of spatial properties such as object position or size, we \textit{neutralize} the slots by performing slot manipulation with the instructions for moving the objects to the central position and for scaling the objects to the unified size. 
After neutralization, we generate a \textit{object image} by decoding a neutralized slot.
The relevance scores between the target object image and candidate object images are computed along the given metric, specifically the MSE.
Finally, object-centric image retrieval can be accomplished by finding the image containing the top-k objects as shown in Fig. \ref{fig:retrieval}.

\subsection{Additional t-SNE Results}
\begin{figure}
    \centering
    \includegraphics[width=\linewidth]{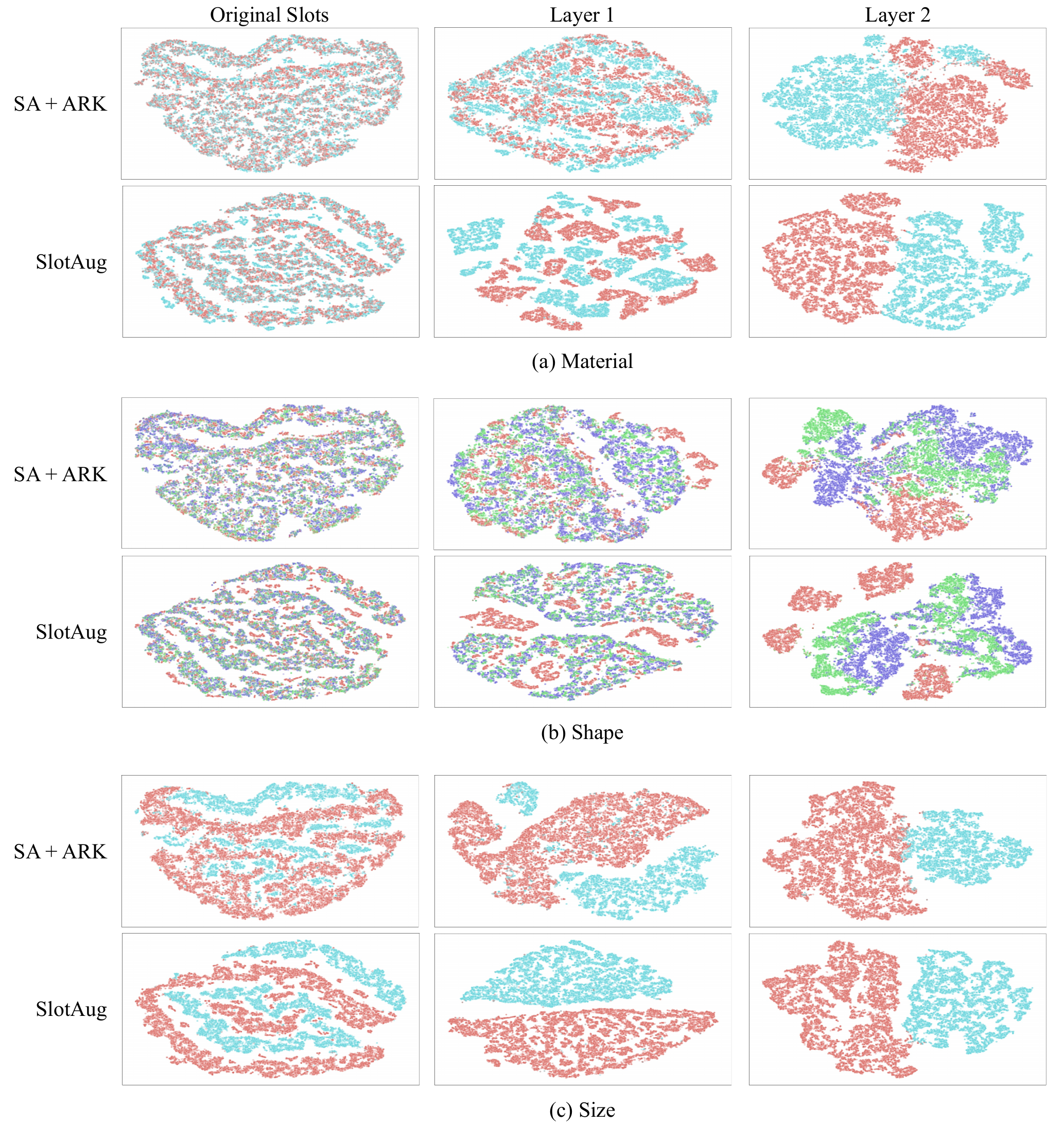}
    \vspace{-1em}
    \caption{ 
    \textbf{The further t-SNE results} from the property prediction.
    We visualize t-SNE for three additional properties: (a) materials (2 types), (b) shape (3 types), and size (2 types).
    }
    \label{fig:additional_tsne}
\end{figure}

Additional t-SNE results from the property prediction are shown in Fig. \ref{fig:additional_tsne}. 
Similar to the result on the color property in the main paper, we can observe that the proposed SlotAug produces more well-clustered slots in the earlier layer in the property predictors.
 
\phantom{0}


\label{subsec:retrieval}

\twocolumn
\bibliography{icml2024}
\bibliographystyle{icml2024}


\end{document}